%% file: main_and_supp.tex
\documentclass[runningheads]{llncs}
\usepackage{graphicx}
\usepackage[numbers,sort]{natbib}
\usepackage{tikz}
\usepackage{amsmath,amssymb}
\usepackage{array}
\usepackage{bbding}
\usepackage{booktabs}
\usepackage{capt-of}
\usepackage{color, soul}
\usepackage{comment}
\usepackage{cuted}
\usepackage{lipsum}
\usepackage{multicol}
\usepackage{multirow}
\usepackage{pifont}
\usepackage{wasysym}
\usepackage{xcolor}
\usepackage{xspace}
\usepackage{cleveref}
\usepackage{url}

\usepackage{appendix}
\usepackage{titletoc}

\input{vedaldi}

\newcommand\norm[1]{\lVert#1\rVert}
\newcommand\method{E2EVE\xspace}
\newcommand\posaugment{\texttt{pos\_augment}\xspace}
\newcommand\sizeaugment{\texttt{size\_augment}\xspace}

\usepackage[accsupp]{axessibility}  % Improves PDF readability for those with disabilities.
\definecolor{MidnightBlue}{rgb}{0.1, 0.1, 0.44}
\definecolor{LimeGreen}{rgb}{0.2, 0.8, 0.2}
\definecolor{Pumpkin}{rgb}{1.0, 0.46, 0.09}
\definecolor{Red}{rgb}{0.8, 0.0, 0.0}
\definecolor{Black}{rgb}{0.0, 0.0, 0.0}
\definecolor{Blue}{rgb}{0.0, 0.0, 0.8}
\definecolor{lavenderblue}{rgb}{0.9, 0.9, 1.0}
\definecolor{mistyrose}{rgb}{1.0, 0.89, 0.88}

\newcommand{\AB}[1]{{\textcolor{Black}{#1}}}

\providecommand{\etal}[0]{\emph{et al.}}

\definecolor{light}{rgb}{0.5, 0.5, 0.5}
\def\light#1{{\color{light}#1}}

\newcommand{\xmark}{\ding{55}}
\begin{document}
% \renewcommand\thelinenumber{\color[rgb]{0.2,0.5,0.8}\normalfont\sffamily\scriptsize\arabic{linenumber}\color[rgb]{0,0,0}}
% \renewcommand\makeLineNumber {\hss\thelinenumber\ \hspace{6mm} \rlap{\hskip\textwidth\ \hspace{6.5mm}\thelinenumber}}
% \linenumbers
\pagestyle{headings}
\mainmatter
\def\ECCVSubNumber{841}  % Insert your submission number here

\title{End-to-End Visual Editing with a \\ Generatively Pre-Trained Artist}

% INITIAL SUBMISSION 
\begin{comment}
\titlerunning{ECCV-22 submission ID \ECCVSubNumber} 
\authorrunning{ECCV-22 submission ID \ECCVSubNumber} 
\author{Anonymous ECCV submission}
\institute{Paper ID \ECCVSubNumber}
\end{comment}
%******************

% CAMERA READY SUBMISSION
%\begin{comment}
\titlerunning{End-to-End Visual Editing with a Generatively Pre-Trained Artist}
% If the paper title is too long for the running head, you can set
% an abbreviated paper title here
%
\author{Andrew Brown\inst{1,2}\and
Cheng-Yang Fu\inst{2} \and
Omkar Parkhi\inst{2} \and \\
Tamara L. Berg\inst{2} \and
Andrea Vedaldi\inst{1,2}}
\authorrunning{Brown et al.}
% First names are abbreviated in the running head.
% If there are more than two authors, 'et al.' is used.
%
\institute{
Visual Geometry Group, University of Oxford \email{\{abrown\}@robots.ox.ac.uk}  \and 
Meta AI
\email{\{chengyangfu, omkar, tlberg, vedaldi\}@fb.com}
\url{https://www.robots.ox.ac.uk/~abrown/E2EVE/}}

%\end{comment}
%******************
\maketitle

\begin{figure}
\centering
\includegraphics[width=\textwidth]{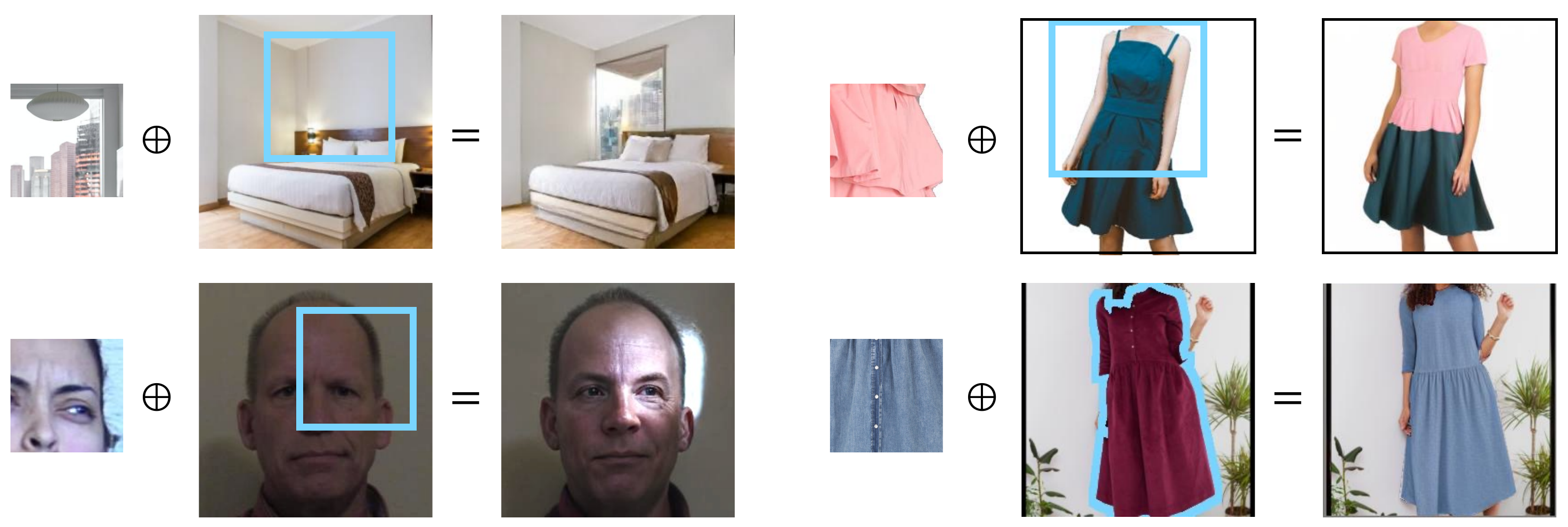}
\caption{\small{\method combines a driver and source image (resp.~to the left and right of the $\oplus$ symbol), generating a new version of the source that resembles the driver in the edit region (marked in blue).
The generated output looks realistic while faithfully resembling the driver.
Our method can be trained to work well on different types of images, including bedrooms, dresses, and faces, and can use regions of arbitrary shape, from rough rectangles to pixel-accurate segmentations (bottom-right).}}\label{fig:teaser}
\end{figure}
\vspace{-10mm}
% \begin{figure}
% \centering
% \includegraphics[height=6.5cm]{eijkel2}
% \caption{One kernel at $x_s$ ({\it dotted kernel}) or two kernels at
% $x_i$ and $x_j$ ({\it left and right}) lead to the same summed estimate
% at $x_s$. This shows a figure consisting of different types of
% lines. Elements of the figure described in the caption should be set in
% italics,
% in parentheses, as shown in this sample caption. The last
% sentence of a figure caption should generally end without a full stop}
% \label{fig:example}
% \end{figure}

\begin{abstract}
\input{Sections/abstract}
% \todo{pass to include end-to-end and method name} 
\end{abstract}

\section{Introduction}%
\label{introduction}
\input{Sections/Introduction}

\section{Related Work}%
\label{Related_Work}
\input{Sections/related_work}

\section{Method}%
\label{Method}

\input{Sections/method}

\section{Experiments}%
\label{Experiments}
\input{Sections/experiments}

\section{Conclusions, Limitations and Future work}%
\label{Conclusion}
\input{Sections/conclusions}

\clearpage

\newcommand*{\toccontents}{\@starttoc{toc}}

\newcommand{\beginsupplement}{%
        \setcounter{table}{0}
        \renewcommand{\thetable}{S\arabic{table}}%
        \setcounter{figure}{0}
        \renewcommand{\thefigure}{S\arabic{figure}}%
     }
\setcounter{tocdepth}{2}
        \makeatletter
        \renewcommand*\l@author[2]{}
        \renewcommand*\l@title[2]{}
        \makeatletter
        
\title{End-to-End Visual Editing with a \\ Generatively Pre-Trained Artist \\ Supplementary Material } % Replace with your title

\titlerunning{End-to-End Visual Editing with a Generatively Pre-Trained Artist}
% If the paper title is too long for the running head, you can set
% an abbreviated paper title here
%
\author{Andrew Brown\inst{1,2}\and
Cheng-Yang Fu\inst{2} \and
Omkar Parkhi\inst{2} \and \\
Tamara L. Berg\inst{2} \and
Andrea Vedaldi\inst{1,2}}
\authorrunning{Brown et al.}
% First names are abbreviated in the running head.
% If there are more than two authors, 'et al.' is used.
%
\institute{
Visual Geometry Group, University of Oxford \email{\{abrown\}@robots.ox.ac.uk}  \and 
Meta AI
\email{\{chengyangfu, omkar, tlberg, vedaldi\}@fb.com
}}
\maketitle
% \setcounter{page}{1}
% \DoToC
{\let\clearpage\relax \tableofcontents}
\renewcommand\thesection{\Alph{section}}
\section{Additional Qualitative Results}%
\label{Additional_Qual}
\input{Supp/Qualitative_Results}

\newpage

\section{Additional Method Details}%
\label{additional_method_details}
\input{Supp/Implementation_Details}

\section{Additional Baseline Details}%
\label{additional_baseline_details}
\input{Supp/Baseline_Implementation_Details}

\section{Additional Quantitative Results}%
\label{additional_quantitative_results}
\input{Supp/Quantitative_Results}

\clearpage

\vspace{5mm}
\bibliographystyle{splncs04}
\bibliography{refs,egbib}
\end{document}

%% file: vedaldi.tex
\newif\ifdark
\IfFileExists{/Users/vedaldi/.bash_profile}{
\immediate\write18{%
if defaults read -g AppleInterfaceStyle 2>/dev/null;
then echo \\darktrue > /tmp/displaymode.tex;
else echo \\darkfalse > /tmp/displaymode.tex; fi}
\input{/tmp/displaymode.tex}
\ifdark
\definecolor{pcolor}{HTML}{1E1E1E}
\definecolor{tcolor}{HTML}{C5C5C5}
\else
\definecolor{pcolor}{HTML}{FDF6E3}
\definecolor{tcolor}{HTML}{333333}
\fi
\pagecolor{pcolor}
\color{tcolor}
\hbadness=\maxdimen
\vbadness=\maxdimen
\vfuzz=30pt
\hfuzz=30pt
}{} 

%% file: Sections/abstract.tex
  We consider the targeted image editing problem: blending a region in a source image with a driver image that specifies the desired change.
Differently from prior works, we solve this problem by learning a conditional probability distribution of the edits, \textit{end-to-end}.
Training such a model requires addressing a fundamental technical challenge: the lack of example edits for training.
To this end, we propose a self-supervised approach that simulates edits by augmenting off-the-shelf images in a target domain.
The benefits are remarkable:
implemented as a state-of-the-art auto-regressive transformer, our approach is simple, sidesteps difficulties with previous methods based on GAN-like priors, obtains significantly better edits, and is efficient.
Furthermore, we show that different blending effects can be learned by an intuitive control of the augmentation process, with no other changes required to the model architecture.
We demonstrate the superiority of this approach across several datasets in extensive quantitative and qualitative experiments, including human studies, significantly outperforming prior work. 

%% file: Sections/Introduction.tex
A key part of the creative process is the ability to combine known factors in novel ways. For instance, we can imagine how a dress would look like with a different v-neck, or our bedroom would look like with the large windows we have seen in a magazine. In this paper, we thus consider the problem of generating new variants of a source image, guided by another image containing a feature, such as a component of a dress or window style, that we wish to change in the source. For additional control, we wish the edit operation to focus on a particular target region of the source, leaving the context as unchanged as possible (see~\cref{fig:teaser}).

Prior works consider image editing tasks, but often guided by a \emph{textual} description of the desired change~\cite{DALLE_2,GLIDE,make_a_scene}.
We argue that specifying edits visually rather than textually offers a far more fine grained and explicit level of control, ultimately resulting in a more useful editor\footnote{After all, a picture is worth a thousand words!}.
Formally, we can describe the editing process as drawing a sample from a conditional image distribution $P(\hat x|x,y,R)$, where $x$ is the source image, $y$ is the driver image, $R$ the edit region and $\hat x$ is an updated version of the source $x$.
The goal of the edit $\hat x$ is to look natural while being close to the source $x$ everywhere except for the region $R$, where it should resemble the driver image $y$.

A main challenge in learning the model $P(\hat x|x,y,R)$ is the lack of suitable training data, namely quadruplets $(\hat x, x, y, R)$, that exemplify the desired mapping.
Most authors have thus proposed to focus on learning an unconditional prior distribution $P(x)$ on images, for which abundant training data is usually available, and then seeking an edit $\hat x$ that is both likely according to the prior and close in some sense to the driver $y$.
This can be achieved in a pre-processing~\cite{zhu2020indomain,chai2021latent} or post-processing stage~\cite{bau2021paint}, and often uses a Generative Adversarial Network (GAN) to model the prior $P(x)$.
Although demonstrating some impressive results, such approaches offer limited control on the edit $\hat x$, which either shows only a weak dependency on the driver image $y$, or does not stay on the image manifold $P(x)$, resulting in undesirable artifacts.

% However there is a significant challenge when training models for the task of semantic manipulation. Namely, there exists no paired data in sufficient scale for images and their edited counterparts, meaning models cannot be trained end-to-end for this task. Instead, current approaches use off-the-shelf unconditional generative models as either a pre-processing (cite) or post-processing stage (cite) in a semantic image-editing pipeline. Although demonstrating some impressive results, the disjoint pipeline means that there is little constraint for the resulting image to remain on the real-image manifold learnt by the generative model. This challenge is understandable, as paired data would be very expensive to obtain, requiring a human image-editing expert potentially many hours for each sample. 

In this work, we overcome these challenges by considering a different approach where we learn the conditional distribution $P(\hat x|x,y,R)$ directly, \emph{end-to-end}.
For this, we propose new ways of \emph{synthesising} suitable training quadruplets $(\hat x,x,y,R)$ on a large scale and without requiring manual intervention.
We do this in a self-supervised manner: given an image $\hat x$, we select an edit region $R$ at random and use it to decompose the image into source $x$ and driver $y$ images, so that the edit can be written as a (random) function $(x,y,R) = f(\hat x)$ of $\hat x$.
A shortcoming is that such $x$ and $y$ are statistically correlated, whereas in a ``creative'' edit process a user must be able to choose $y$ independently of $x$.
A key contribution is to show that, if the process $f$ is carefully designed, then the resulting images $x$ and $y$ are independent \emph{enough}, meaning that they can be used to learn a high-quality conditional generator $P(\hat x | x, y,R)$ which works even when $x$ and $y$ are sampled independently.

We pair this intuition with the adoption of state-of-the-art auto-regressive image modelling using transformers for learning and sampling the conditional distribution $P(\hat x | x, y, R)$.
Overall, our End-to-End Visual Editor (\method) approach has significant advantages over prior image editing works:
(1) based on extensive qualitative, quantitative and human-analysis experiments on several datasets, it results in higher quality edits that are simultaneously more dependent on the driver image and more natural looking than prior works based on GANs and attention;
(2) it is generally easier to implement and tune than GAN-based alternatives; and 
(3) it is efficient because it allows to sample directly edits without involving expensive pre- or post-processing steps required by some prior methods.
%require to optimize the output image iteratively.
Code implementing our models will be made available on publication.

%% file: Sections/related_work.tex
\paragraph{Targeted Generative Image Editing. }

% In this work, we are focused on manipulating images in certain targeted regions using generative models, whilst keeping the rest of the image largely unchanged. 

Approaches for editing images in targeted locations include spatial manipulation of objects~\cite{Zhu2016GenerativeVM}, adding or removing a closed-set of objects~\cite{bau2020units,sem_photo_manip}, or text-driven manipulation using CLIP~\cite{bau2021paint,shi2021spaceedit}.
These GAN-based approaches have some downsides: First, they require inverting GANs to represent the input image --- a difficult problem~\cite{Im2SG,Abdal2019Image2StyleGANHT,Zhu2016GenerativeVM,Lipton2017PreciseRO,bau2019inverting,Guan2020CollaborativeLF} which can limit the \textit{editability} of the images~\cite{tov2021designing,Im2SG}.
Second, they are not trained end-to-end.
Third, text-driven approaches offer limited fine-grained control of shape and texture~\cite{bau2021paint}.
We address such shortcomings by training a sequence-to-sequence model end-to-end for the task of targeted image-based visual editing.

% In this work we instead focus on visual-driven image manipulation, which offers users far more fine-grained and explicit control of the editing process.
% By training sequence-to-sequence generators, we negate the need for image-inversion, and by training a model end-to-end for the editing task, the output images from our model are more dependant on the driver, and are more realistic.

% Note that targeted image manipulation, which we address in this paper, differs from global image manipulation~\cite{Radford2016UnsupervisedRL,Karras2019ASG,shen2020interpreting,gansteerability,yang2019semantic,choi2018stargan,Plumerault2020ControllingGM,style_flow,Schwettmann_2021_ICCV,Hrknen2020GANSpaceDI,voynov2020unsupervised,peebdisent,Wu2020StyleSpaceAD,wang2021aGANGeom,shen2020interpreting,yksel2021latentclr,voynov2020unsupervised,ZhuangICLR2021,patashnik2021styleclip,testwebsite3,testwebsite4,generating2021,styleclipdraw,xu2021predict,kwon2021clipstyler,liu2021control,abdal2021clip2stylegan,xia2021tedigan} and spatially-conditioned generation~\cite{pix2pix2017,Park_2019_CVPR,Wang2018HighResolutionIS,kim2021stylemapgan,xia2021tedigan,huang2021multimodal,zhao2021comodgan}, where an entire image is generated/manipulated.
Note that the targeted image manipulation task, \AB{which we address in this paper differs from spatially-conditioned generation~\cite{pix2pix2017,Park_2019_CVPR,Wang2018HighResolutionIS,kim2021stylemapgan,xia2021tedigan,huang2021multimodal,zhao2021comodgan}, or global image manipulation, where an entire image is generated/manipulated~\cite{Radford2016UnsupervisedRL,Karras2019ASG}. In global manipulation works, interpolations in the latent space are located which correspond to edits over the entire image, either via visual attribute classifiers~\cite{shen2020interpreting,gansteerability,yang2019semantic,counterfactual,choi2018stargan}, unsupervised disentanglement~\cite{Plumerault2020ControllingGM,style_flow,Schwettmann_2021_ICCV,Hrknen2020GANSpaceDI,voynov2020unsupervised,peebdisent,Wu2020StyleSpaceAD,wang2021aGANGeom,shen2020interpreting,yksel2021latentclr,voynov2020unsupervised,ZhuangICLR2021}, or via image-text similarity~\cite{patashnik2021styleclip,testwebsite3,testwebsite4,generating2021,styleclipdraw,xu2021predict,kwon2021clipstyler,liu2021control,abdal2021clip2stylegan,xia2021tedigan}.}
Our task is also different to image in-painting~\cite{inpaint1,liu2018partialinpainting,peng2021generating,yu2018generative,wan2021high,bert_inpaint}, where the generation of image regions is dependent only on the surrounding context.
In our task, the generation further depends on a driver image.

\paragraph{Image Composition.}

Image composition combines different and possibly inconsistent images into a single realistic and cohesive output.
Previous approaches include image collaging using nearest neighbors~\cite{SceneCollaging}, using auto-encoders to compose foreground and background~\cite{Tsai2017DeepIH}, composing a closed set of visual attributes~\cite{mask_collate,press2018emerging},  or using semantic pyramids~\cite{shocher2020semantic}.
Others fuse images by projecting their composition on the manifold generated by a GAN via inversion~\cite{xu2021generative,chai2021latent,ghosh2021invgan,Im2SG,zhu2020indomain,issenhuth2021edibert}.
% Recent approaches have demonstrated the ability of the prior learnt by an unconditional generative model to project manually composited images into cohesive realistic-looking outputs, either just by GAN-inversion~\cite{xu2021generative,chai2021latent,ghosh2021invgan}, or by an iterative process~\cite{Im2SG,zhu2020indomain,issenhuth2021edibert}.
These methods work well if the driver image is sufficiently aligned to the source (which usually requires manual intervention), but worse than our end-to-end model when this is not the case.

\paragraph{Two Stage Image Synthesis.}

We adopt state-of-the-art two stage auto-regressive models~\cite{dai2018diagnosing,Esser2020ADI,Xiao2019GenerativeLF,image_gpt,Ramesh2021ZeroShotTG,Esser2021TamingTF,ding2021cogview} for the image generator.
These models scale better than sequence-to-sequence~\cite{radford2019language,Devlin2019BERTPO} models applied directly to pixels by reducing first the dimensionality of images via a discrete autoencoder~\cite{vqvae,vq_vae2,Fauw2019HierarchicalAI,Esser2021TamingTF}.
% To alleviate the computational cost of training sequence-to-sequence~\cite{radford2019language,Devlin2019BERTPO} models directly on pixels, the two-stage approach learns a condensed discrete image representation~\cite{vqvae,vq_vae2,Fauw2019HierarchicalAI,Esser2021TamingTF} first, on which the generator is trained.
% In particular, we follow Esser~\textit{et al.}~\cite{Esser2021TamingTF} that trained impressive image-conditional models and introduced the VQ-GAN for the first stage.
% Recent works have built upon the approach of~\cite{Esser2021TamingTF} fo
Others have recently built on this work for text- or class-driven image manipulation~\cite{testwebsite4,wu2021n,esser2021imagebart}, whereas we consider image-driven editing.
% Different to these works, we utilise two stage image synthesis for visual-driven targeted image editing.

{}\cite{zhang2021ufcbert} shows a (single) qualitative result for image-based out-painting using a two-stage BERT~\cite{Devlin2019BERTPO} model, where a small set of tokens from the output sequence is fixed (termed \textit{preservation controls}), and paired training data is sourced from the same image.
We do not solve the out-painting problem, but the target editing problem, with the added challenge of preserving context while mixing images with very different statistics.

%% file: Sections/method.tex
% \begin{comment}
% \begin{figure}[t!]
%   \centering
%   \includegraphics[width=0.9\linewidth]{figures/info_restriction_v2.pdf}
%   \caption{Left: A graph showing the negative log likelihood (NLL) on the training set from 5 different \method models trained with varying amounts of information restricted from the model while training with paired inputs. Middle: An example of non-paired inputs during inference, where the masked context image $\dot{x}_{R}$, and visual cue, $\tilde{v}$ are taken from different images. Right: Samples taken from three of the models shown on the graph on the left when conditioned on the non-paired inputs shown in the middle. The goal of realistic images that depend on the visual cue can be obtained via the correct balance of information restriction. }
%   \label{fig:cut_out_regions}
% \end{figure}
% \end{comment}

We wish to learn a model that can ``naturally'' blend a given source image $x$ with a user-provided driver image $y$.
Formally, we denote with $x,\hat x \in \mathbb{R}^{3\times H \times W}$ the source and output (RGB) images and with $y \in \mathbb{R}^{3 \times H'\times W'}$ the driver image (where, usually, $H > H'$ and $W > W'$).
Furthermore, we target the edit operation on a region  $R \in \{0,1\}^{H \times W}$, expressed as a binary mask.
We cast the problem as one of learning a conditional probability distribution $P(\hat x|x,y,R)$ and then sample the output image $\hat x$ conditioned on the source image $x$, the driver image $y$, and the edit region $R$ (\cref{fig:teaser}).

\begin{figure}[t!]
  \centering
  \includegraphics[width=0.88\linewidth]{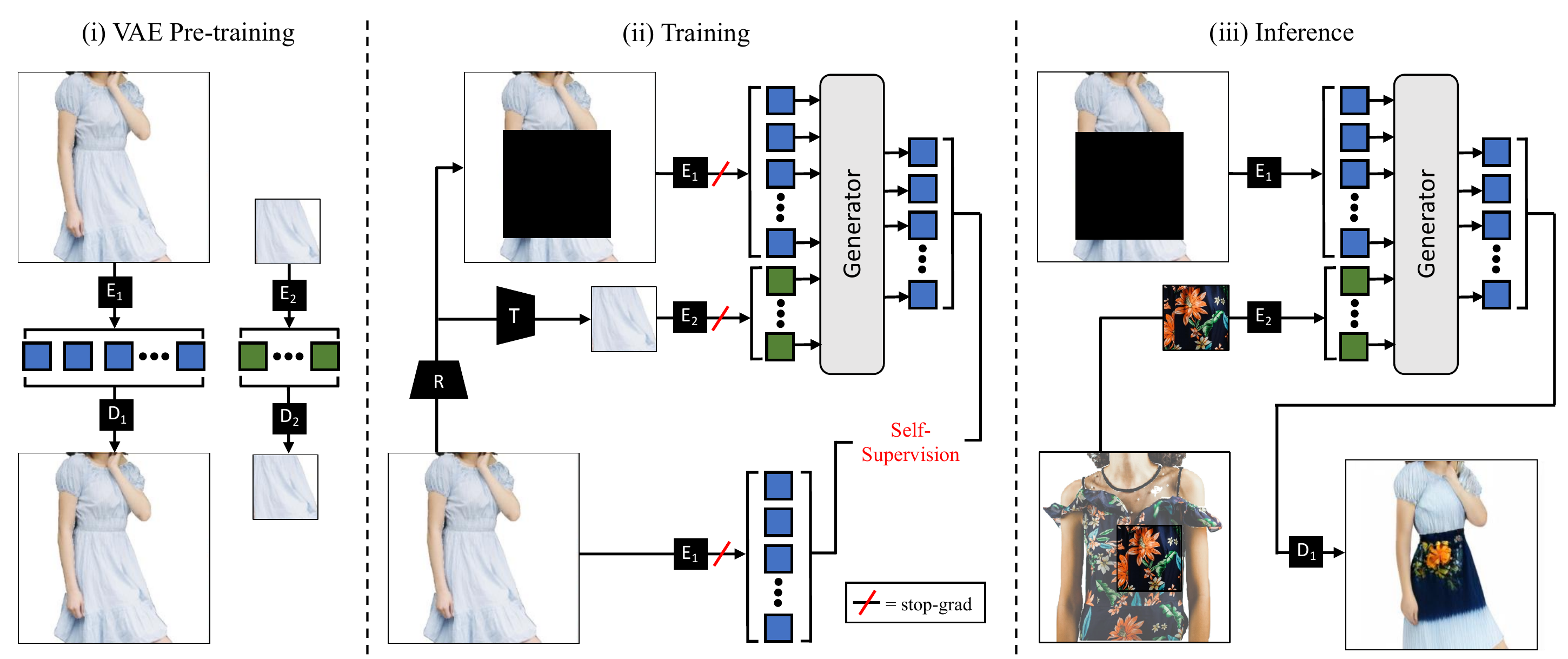}
  \vspace{-3mm}
   \caption{\footnotesize{The \method approach: (i) \textbf{VAE Pre-Training}: we train two quantized VAEs (one for whole images, and one for image patches), each consisting of an Encoder, $E$, and decoder, $D$. (ii) \textbf{Training}: Each data sample produces a masked source image (via the operation $R$), and driver image (via random transformation $T$ on the masked region). Given these conditioning inputs, the model is self-supervised to predict the data sample. Following prior work, the VAEs are kept frozen while training the generator. (iii) \textbf{Inference}: \method generates edited images when the source and driver are sampled independently from different images.}\label{fig:method}}
  \vspace{-4mm}
\end{figure}

Next, we discuss the advantages and requirements of this approach (\cref{s:direct}), propose a self-supervised learning formulation that does not require any manually-provided labels to train the model (\cref{s:data}), and give the technical details of the neural network that learns the conditional distribution (\cref{AR_models}).
An overview of our training and inference settings is shown in \cref{fig:method}.

\subsection{End-to-End Conditional Generation}\label{s:direct}

Our approach is to learn a conditional distribution $P(\hat x | x, y, R)$ \emph{end-to-end}, which we do by means of an auto-regressive transformer network discussed in~\cref{AR_models}.
In order to train such a model, we require training quadruplets $(\hat x, x, y, R)$ sampled from the joint distribution $P(\hat x, x, y, R)$.
Each of these quadruplets represents the outcome of a ``creative'' process, where a human artist combines images $x$ and $y$ to generate a new image $\hat x$.
Because obtaining such training data would require the intervention of human artists, it would be very difficult to obtain a sufficiently large dataset to learn the required conditional distribution.
Hence, much of the research in image editing focuses on how to avoid this bottleneck and use instead data which is readily available.

A popular approach is to consider an \emph{indirect} formulation and learn instead an unconditional image distribution $P(x)$, for instance expressed as a GAN generator $x = G(z)$.
Then, the output image $\hat x = G(z^*)$ is ``sampled'' via an optimization process like
$
z^* = \operatornamewithlimits{argmin}_z d(G(z)|_R, y)
$
where $d(\hat x|_R, y)$ measures compatibility between the region $R$ of the generated image $\hat x$ and the driver image $y$.
The advantage is that the model $G$ can be learned from a collection $\mathcal{X}$ of unedited images $x \sim P(x)$, which is often easy to obtain at scale.
The disadvantage is that this model is not optimized for the final task of image editing.
% and the inference process is relatively expensive as it entails solving an optimization problem.

By contrast, in our approach we learn directly the model $P_\theta(\hat x | x, y, R)$ minimizing the standard negative log-likelihood loss:
\begin{equation}\label{e:logloss}
\theta^*= \operatornamewithlimits{argmin}_{\theta}
\left(
- \frac{1}{|\mathcal{T}|}
\sum_{(\hat x, x, y, R) \in \mathcal{T}} \log P_\theta (\hat x | x, y, R)
\right)
\end{equation}
where $\mathcal{T}$ is a large collection of training quadruplets.
Once learned, we can directly draw samples $\hat x \sim P_{\theta^*}(\hat x | x,y, R)$.
% in a feed-forward manner without further optimization and with better empirical results.
The main challenge is how to obtain the training set $\mathcal{T}$.
The key to our method is a way of constructing $\mathcal{T}$ from $\mathcal{X}$ in an automated fashion, at no extra cost.
This is explained in the next section.

\subsection{Synthesizing a Dataset of Meaningful Edits}\label{s:data}

Given a training set $\mathcal{X}$ of unedited images $x$, the goal is to create a dataset $\mathcal{T}$ of ``edits'' $(\hat x, x, y, R)$ consisting of the generated image $\hat x$, the source image $x$, the diver image $y$, and the edit region $R$.
The difficulty is that these quadruplets should be representative of a ``creative'' process where the generated image $\hat x$ is a meaningful blend of the source and driver images, $x$ and $y$, according to a human artist.
Specifically, $\hat x$ should resemble $x$ as much as possible except in the region $R$, where it should take the character of $y$, but without introducing unnatural artifacts (e.g., simply pasting $y$ on top of $x$ would not do).

\begin{figure}[t!]
  \centering
  \includegraphics[width=\linewidth]{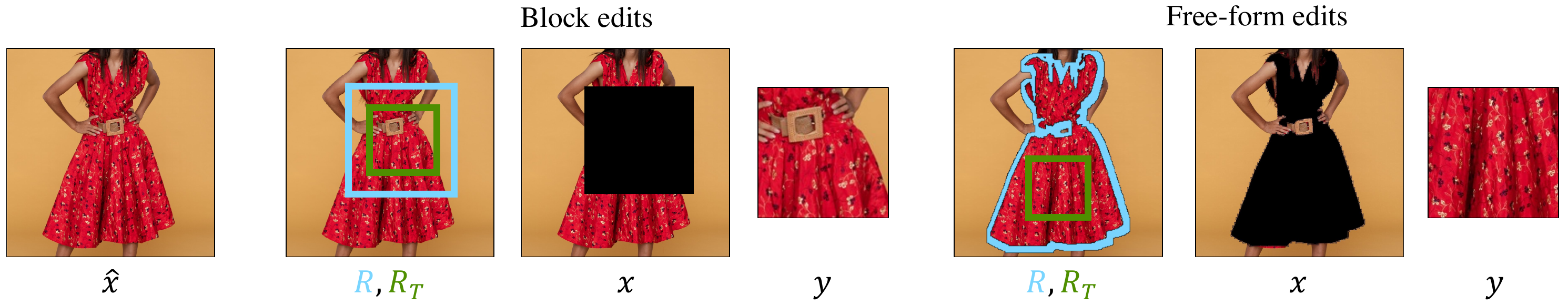}
  \vspace{-6mm}
  \caption{\footnotesize{\emph{Left: block edits.} We sample an output image $\hat x$ and generate the corresponding edit input $(x,y,R)$ by sampling a square region $R$, a transformation $T$ cutting a sub-region $R_T$ from $R$, and extracting the source image $x$ and a driver $y$ from these two.
  \emph{Right: free-form edits.} The same approach, but this time $R$ is a pixel-accurate segmentation mask extracted manually or automatically ( $R_T$ remains a square sub-region) .}}\label{info_restrict_method}
  %  The two methods for constructing the \textit{paired inputs} of the masked context image $\dot{x}_{R}$, and visual cue $\dot{v}$, from the output image $x_{edit}$, while restricting information from the model. Method 1 assumes a square edit region in $R$, and the transformation $\hat{T}$ restricts information by taking a square crop within the masked region. Method 2 assumes that dense spatial annotations are available in the form of semantic segmentations. Here, the edit region $R$ covers the entire spatial extent of an object (in this case, a clothing item), and the transformation $\hat{T}$ takes a square crop from within this region.}
  \vspace{-6mm}
\end{figure}

We propose to build such quadruplets as follows (see also~\cref{info_restrict_method}).
We sample an output image $\hat x$ from the unedited collection $\mathcal{X}$, thus pretending that the latter is, in fact, the result of an edit operation.
Then, we define the source and driver images for this virtual edit as follows:
\begin{equation}
x = (1 - R) \odot \hat x, \qquad
y = T(R \odot \hat x),
\end{equation}
where $R$ is the mask of a random image region, $\odot$ is the element-wise product (where broadcasting is used as required), and $T : \mathbb{R}^{3\times H\times W} \rightarrow \mathbb{R}^{3\times H' \times W'}$ is a \emph{random image transformation}, also known as an ``augmentation''.

By optimizing the log-likelihood loss in~\cref{e:logloss}, the model $P_\theta(\hat x | x, y, R)$ learns to predict the full image $\hat x$ from $x$ (which misses the region $R$), and $y$, which preserves some information about the missing region.
Because $\hat x$ is originally an unedited image, the model learns to predict a natural-looking output.

\begin{figure}[t!]
   \centering
  \includegraphics[width=\linewidth]{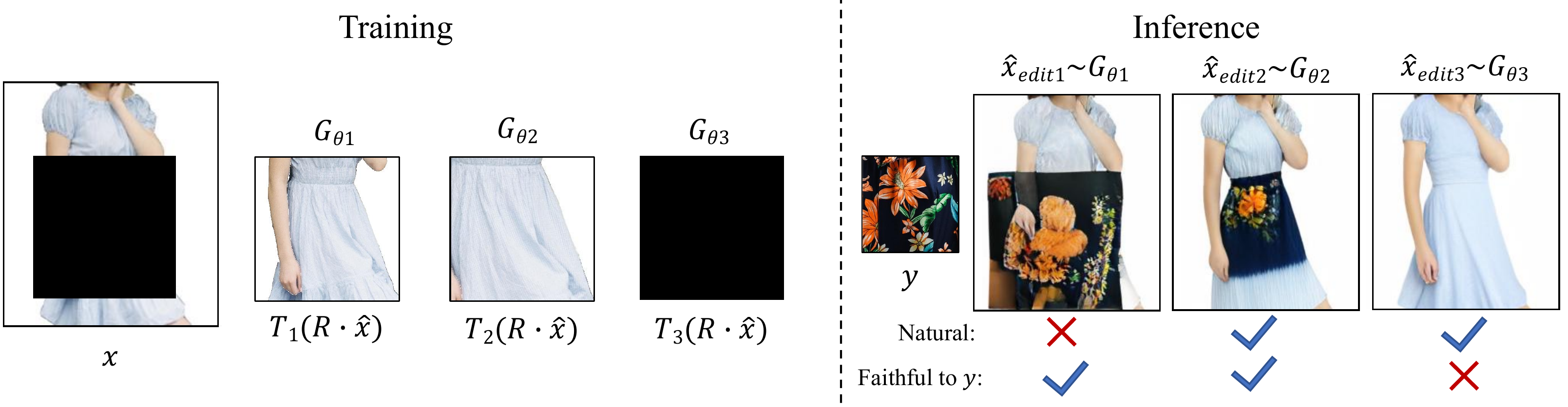}
  \vspace{-6mm}
    \caption{\footnotesize{Intuition for augmenting training inputs. During training, the driver image $y$ is computed by applying a random transformation $T$ to the masked region of the source image $x$. Left: Three options for $T$ that are used to train three different generators, $G_{{\theta}n}$. Right: Samples from the three trained models when conditioned on $x$ and an independently sampled $y$. An optimal choice of $T$ removes just enough information that generated images are both natural-looking, and are faithful to the driver image $y$.} }
    \label{fig:cut_out_regions}
    \vspace{-5.0mm}
\end{figure}

The key design choice in this construction is the random transformation $T$.
For example, if we set $T=1$ to be the identity function, then the output image can be reconstructed exactly as $\hat x = x + y$; in this case, the model $P_\theta(\hat x|x,y,R)$ learns to paste $y$ onto $x$, which is uninteresting (see model $G_{\theta_1}$ in \cref{fig:cut_out_regions}).
On the other hand, if we set $T=0$ to be the null function, then $y$ does not provide any information about the image; in this case, the model  $P_\theta(\hat x|x,y,R)$ learns to inpaint $\hat x$, filling in the missing region in a non-trivial manner, but {ignoring} the driver image $y$ altogether (see model $G_{\theta_3}$ in \cref{fig:cut_out_regions}).
The augmentations $T$ should find a \emph{sweet spot} and remove just the right amount of information from $y$ (see model $G_{\theta_2}$ in \cref{fig:cut_out_regions}).
While finding the optimal choice for the random transformations $T$ is ultimately an empirical process, we describe next some important design criteria that were crucial for our results.

\paragraph{Decorrelating source and driver images.}

A difficulty with our approach is that, because both source image $x$ and driver image $y$ are derived from the same image $\hat x$, they are \emph{not} independent but \emph{paired}.
This is a problem because the user should be free to choose almost \emph{any} driver image $y$ for editing, so the generator must work well for \emph{unpaired} inputs $x$ and $y$ too.
We can approximate this condition by making $x$ and $y$ as uncorrelated as possible during training.
% Intuitively, this can be achieved by focusing $x$ and $y$ on different \emph{semantic components} of the image $\hat x$ (for example, window vs.~bed in a bedroom) because they should naturally be more independent.

While we cannot achieve this result exactly, we propose two methods to approximate it, \textit{block} and \textit{free-form}, both shown in~\cref{info_restrict_method}.
\textbf{Block edits} are simple: we let the transformation $T(R \odot \hat x)$ take a further sub-crop $R_T$ of the crop $R$ it is given as input, thus removing the most direct source of correlation between $x$ and $y$: spatial continuity.
Furthermore, we found empirically that if the sub-crop is always centered in $R$ and of a fixed relative size w.r.t. $R$, then the model learns, as one would expect, to paste this crop in the middle.
Instead, we let $T$ further randomize the position (\posaugment) and size (\sizeaugment) of the sub-crop relative to the edit region during training. We parameterize the sub-cropping operation via $\alpha$, which defines the ratio of the sub-crop width, to the edit region width. The \sizeaugment operation allows $\alpha$ to vary during training.
The model hence learns to find a meaningful placement for the patch $y$ in the context of $x$, without assuming spatial continuity or a specific geometric arrangement.
Optionally, we further decorrelate source and driver images via \textbf{free-form edits}.
The difference is that we let $R$ be the output of a semantic segmentation network, while $T$ still takes a square sub-crop $R_T$ from region $R$.
Because $y$ is fully contained in the edit region $R$ and the latter separates a foreground object from the background, this significantly reduces the correlation between $x$ and $y$.
While this approach requires additional machinery (e.g., a segmentation network), empirically it can obtain impressive results (\cref{big_results}).

\paragraph{Controlling the learned editor.}

In order to favour generalization, the augmentations above should remove as much information as possible from the crop $y$ \emph{except} for the information that the editor should transfer from the driver image $y$ to the generated image $\hat x$.
For example, it would be possible to consider further augmentations such as color jitter, but this would cause the editor to learn to ignore the color, which we usually wish to transfer.
In general, by choosing different augmentations we can \emph{control} what information the editor learns to transfer from the driver image to the generated one (e.g., style and colour), and what to ignore (e.g., the specific spatial arrangement).

\vspace{-2mm}
\subsection{Two-Stage Conditional Auto-Regressive Image Generation}\label{AR_models}

% In this section, we motivate the choice of auto-regressive models for the generator, $G$, and provide some theory for their application to image generation. We model the distribution in Equation~\ref{prob_def} using an auto-regressive (AR) model. 

In order to implement the conditional distribution $P_\theta(\hat x | x, y, R)$, we use an \emph{auto-regressive (AR) model}.
AR models have been shown to be highly expressive for image generation~\cite{Ramesh2021ZeroShotTG,Esser2021TamingTF}, they can be conditioned on multiple signals elegantly and without architectural changes, and, unlike GANs~\cite{salimans2016improved}, are mode-covering.
In practice, this leads to more varied generation results and the ability to model datasets with more variation.
We summarise next how this model is applied to our case and point the reader to the supp.~mat.~for additional details.

The goal is to model a conditional distribution $P(\hat x | c)$, where $c$ lumps together all conditioning information.
An AR model further decomposes $\hat x = (\hat x_1,\dots,\hat x_M)$ into $M$ components and factorizes the distribution as the product
% \begin{equation}\label{auto-regressive}
$
P(\hat x|c) = \prod_{m=1}^{M} P(\hat x_m | \hat x_1, \dots, \hat x_{m-1}, c).
$
% \end{equation}
The model is trained by minimizing the negative log likelihood~\eqref{e:logloss} (avoiding unstable adversarial techniques used in GANs). For modelling images $x$, a challenge lies in finding a suitable decomposition, such that the individual factors $P(\hat x_m | \hat x_1, \dots, \hat x_{m-1}, c)$ can be implemented effectively.
To this end, we build upon the two stage process of Esser~\etal~\cite{Esser2021TamingTF} and use a transformer on top of a discrete autoencoder. Note that the focus of this work is on end-to-end targeted image editing and the training formulation; we describe the two-stage auto-regressive method for completeness and reproducibility.

Specifically, in the first stage we learn a compressed and discretized representation $z=\Phi(\hat x) \in \{1,\dots,K\}^M$ of the images, where here $K$ is the size of the discrete encoding space and $M$ the resulting number of discrete tokens.
For this, we use the VQ-GAN method of~\cite{Esser2021TamingTF}, achieving a 16-fold compression of the images (we use separate encoders for $x$ and $y$).
Naturally, the encoder comes with a paired decoder $\hat x = \Psi(z)$ that allows to reconstruct the image from the code.
This achieves two important goals: (1) it allows to scale the generator model to higher resolution images, which is important for visual quality; and (2) it allows, via discretization, to predict discrete distributions for the second stage.

The second stage uses a transformer to model the factors $P(\hat x_m | \hat x_1, \dots, \hat x_{m-1}, c)$.
%in~\cref{auto-regressive}.
Specifically, recall that the conditioning information $c=(x,y,R)$ consists of the source image $x$, the driver image $y$ and the region $R$.
The sequence of tokens $S_m = (z_1,\dots,z_m) \oplus \Phi(x) \oplus \Phi(y)$ (where $\oplus$ denotes concatenation), comprising the partially-predicted output tokens along with the conditioning tokens, is fed to the transformer to output a $K$-dimensional histogram $P(z_m = \cdot | S_m)$.
Spatial encodings are added to the image tokens, but there is no need to explicitly encode the region $R$ as the latter can be inferred from $x$ because $x = (1 - R)\odot \hat x$ has a $R$-shaped `hole'.
While this way of encoding for $R$ may seem na{\"{i}}ve, it is in fact simple and powerful:
prior work such as EdiBERT~\cite{issenhuth2021edibert} use ``occlusion tokens'', and hence lose the ability to express pixel-accurate edit regions, which we can do effortlessly.

As for the model details, we train a GPT-2~\cite{radford2019language} style transformer.
The factors $P(\hat x_m | \hat x_1, \dots, \hat x_{m-1}, c)$
%  in \cref{auto-regressive}
require each predicted token to depend only on those prior to it in the sequence. Hence,
GPT-2 uses causal masking allowing only unidirectional attention towards earlier tokens in the sequence. All factors are trained efficiently in parallel using \textit{teacher forcing}~\cite{teacher_forcing,prof_forcing}. During inference, only the conditioning information is provided so the target sequence is predicted iteratively, sampling one symbol $z_m$ at a time from the corresponding histogram.
In practice, inference is faster than for some GAN alternatives, as shown in the sup.~mat.

%% file: Sections/experiments.tex
We compare our method to others that, given a source image $x$ and a driver image $y$, produce one or more edits $\hat x$.
Good edits have three properties:
(1) \emph{naturalness} (the edit $\hat x$ looks like a sample from the prior $P(x)$);
(2) \emph{locality} ($\hat x$ is close to $x$ outside the edit region $R$ --- although a certain amount of slack is necessary to allow the edit to blend in naturally);
and
(3) \emph{faithfulness} ($\hat x$ resembles $y$ within the edit region $R$).
Achieving only one of the three objectives is trivial (for example, setting $\hat x = x$ ignoring $y$ is natural and local but unfaithful whereas copying $y$ on top of $x$ is is local and faithful but unnatural) so a good model must seek for a trade-off between these properties. 

Measuring these properties is not entirely trivial; for the quantitative analysis, we take the standard FID measure for naturalness~\cite{FID_paper}, the $L^1$ distance $\|(1-R)\odot (\hat x - x)\|_1$ to measure locality, and a retrieval approach to measure faithfulness.
For the latter, we consider a set $\mathcal{Y}_\text{dstr}$ of 100 distractor images of the same size as the driver image $y$, use the edited region $\hat x|_R$ as a query, and find its nearest neighbour $y^* = \operatornamewithlimits{argmin}_{\hat y \in \{y\} \cup \mathcal{Y}_\text{dstr}} d(\hat x|_R, \hat y)$, incurring the loss $\delta_{y^* \not = y}$.
In this expression, $d(\cdot,\cdot)$ is the Inception v3~\cite{7780677} feature distance (pre-trained on ImageNet~\cite{5206848}) which is the same encoder used for FID calculation.
% Intuitively, the (average of this) loss is low if $\hat x|_R$ allows to easily identify the driver image $y$, meaning that it depends on it.

\paragraph{Evaluation data.}

Recall that our goal is to evaluate the quality of automated editing algorithms.
To feed such algorithms, we need triplets $(x,y,R)$ consisting of a source image $x$, a driver image $y$ and an edit region $R$.
In~\cref{s:data} we explained how to build such a dataset for the purpose of \emph{training} our model --- with the added complexity that, for training, we \emph{also} need to know the result $\hat x$ of the edit process.
We could use the same dataset for evaluation, but this would unfairly advantage our model.
Instead, since knowing the output $\hat x$ is \emph{not} required to measure naturalness, locality and faithfulness, we are free to choose \emph{new} and less constrained triplets $(x,y,R)$ for evaluation, resulting in more challenging edits and a fairer evaluation.
However, we wish to avoid too many cases in which cohesive blending is impossible (\textit{e.g.}, where $y$ is a patch of sky and $x|_R$ is a face).
Hence, we assemble evaluation triplets as follows:
given a sample image $x$ and an edit region $R$, we define $y = x'|_R$ to be a crop taken at the same spatial location from a \emph{different} image $x'$ in the dataset.
The effect is to (very) weakly constrain $x|_R$ and $y$ to be compatible (e.g.,~both sky regions, or face regions) by exploiting the photographer bias in the datasets we consider.

We conduct experiments on three datasets:
(1) the private \textit{Dresses-7m} dataset containing 7 million images mainly depicting a woman wearing a dress;
(2) LSUN bedrooms~\cite{Yu2015LSUNCO} containing  3m images of bedrooms; and
(3) FFHQ~\cite{Karras2019ASG}, containing 70K aligned faces.
We sample $x$ by considering 1024 images from the validation sets of \textit{Dresses-7m} and FFHQ, and 256 for LSUN bedrooms (due to its small size).
For each $(x,y,R)$, we consider 10 edit samples and obtain naturalness, locality and faithfulness by averaging over all images thus generated (totalling 10,240 and 2,560 samples, respectively).
% We evaluate \textit{}Unless otherwise noted, we use basic square crops $R$ (instead of running a semantic segmentation method to obtain $R$). 
\AB{All images in this paper are from UnSplash~\footnote{www.unsplash.com} (dresses and bedrooms), or DFDC~\cite{DFDC2020} (faces).}
% \AB{Details to add regarding Dresses-7m - for the block-edit visualisations, we ``whiten'' the backgrounds using the segmentation model. This vastly improves the results as two conflicting backgrounds confuse the model. For the free-form edits visualisations - I paste the edited region onto the original image - hence the very high quality backgrounds. We cannot do this for the block-edits because the edited region does not align with the non-edited image anymore (the ``refinement'' networks I mention in the limitations section address this issue so that the edited region can be collated back with the original image).}

\paragraph{Implementation Details.}

We use a transformer architecture with 24 layers, 16 head multi-head attention, embedding size 1024, and we train it using standard cross-entropy loss.
The masked source image $(1 - R) \odot x$ and the driver image $y$ are encoded using VQ-GANS with 16$\times$ compression and 1024 codebook size.
Source $x$ and output $\hat x$ images have resolution $256\times256$ and $y$ resolution $64\times64$.
The token sequences $S_m$ (comprising the coded $x$, $y$ and partial $\hat x$) has maximum length 516.
Our final model uses \posaugment and \sizeaugment --- the latter forms $y$ by taking a sub-crop in the region $R$, with $\alpha$ varying from 0.4 to 0.7. We use a batch size of 512, and the AdamW~\cite{Loshchilov2019DecoupledWD} optimizer with learning rate 4.5e-6. During inference, for each input $(x,y,R)$, we generate 20 samples $\hat x$ and keep the 10 with highest similarity to $y$ (a method termed \textit{Filter}, in~\cref{ablation_table}).
In order to focus the sampling on more realistic/likely outputs we use \emph{nucleus sampling}~\cite{Holtzman2020The} with a p-value of 0.9.
Following prior work~\cite{Esser2021TamingTF,Ramesh2021ZeroShotTG}, the VQ-GANs are kept frozen when training the transformer.

% At each auto-regressive step, token samples from the multinomial probability distribution predicted by the model are restricted to the most probable tokens whose cumulative probability is greater than 70\%.

%  is  weights corresponding to the probabilities of the most likely tokens. The most likely tokens are chosen as those having a cumulative probability of greater than 0.7 (\textit{i.e.} top-p sampling with p-value of 0.7).}

%  TODO: Input training details (iterations, batch sizes, learning rates)

% Our goal is to manipulate an image in a certain region, given a visual cue taken from a different image. To this end, we form separate evaluation sets for each dataset, where each instance in the set consists of a context image, an edit region, and a visual cue - a cropped region taken from a separate, target image. A challenge lies in how to select the visual cue for each sample. We are interested in the unconstrained setting where there may be no clear semantic alignment between the visual cue and the edit region of the context image. 

\paragraph{Baselines.}

We compare our method against the following image composition baselines:
(1) \textit{Copy-paste} generates $\hat x$ by pasting $y$ onto $x$ at the specified location $R$;
(2) \textit{Inpaint} ablates our method by removing the tokens $y$ from the input, thus generating $\hat x$ by inpainting the region $R$ unconditionally while disregarding $y$;
(3) \textit{GAN inv}, inspired by~\cite{issenhuth2021edibert,xu2021generative,zhu2020indomain}, takes the copy-paste output and uses the StyleGANv2~\cite{Karras2019stylegan2} or StyleGANv2-ADA~\cite{Karras2020ada} networks to re-encode and thus denoise the resulting image via GAN inversion~\cite{Im2SG}, ``blending'' the edit;
(4) \textit{EdiBERT}~\cite{issenhuth2021edibert} is a related transformer-based approach, which iteratively refines the output of copy-paste output using BERT~\cite{Devlin2019BERTPO} (for fairness, we use the same VQ-GAN and sample filtering by similarity to driver as for our method);
(5) \textit{In-Domain GAN}~\cite{zhu2020indomain} uses a regularised form of GAN inversion to blend source and driver images.
% in-painting model by removing the visual cue conditioning from \method. namely EdiBERT~\cite{issenhuth2021edibert}, and In-Domain GAN~\cite{zhu2020indomain}. Additionally, taking motivation from~\cite{issenhuth2021edibert,xu2021generative,zhu2020indomain} who showed that simply GAN-inverting and decoding a composited image is a strong baseline, we use the technique from~\cite{Im2SG} for inverting composited images produced by the \textit{copy-paste} baseline, and decode the images using StyleGANv2~\cite{Karras2019stylegan2} and StyleGANv2-ADA GANS~\cite{Karras2020ada}.
Pre-trained models are available for all test datasets except \textit{Dresses-7m};
unfortunately, we were unable to successfully train the GAN-based models on the latter (possibly due to the significant diversity of this data), so in this case we limit the other baselines.
% We tune the hyper-parameters and sampling methods of each baseline to get the best performance.
Some off-the-shelf models are trained on the validation sets that we use for testing, which disadvantages our approach in the comparison.
For more details on experimental settings, please see supp.~mat.

% We use pre-trained models for all comparisons, apart from on the \textit{Dresses-7m} dataset where we train an EdiBERT from scratch.
% We were unable to train a GAN successfully on the \textit{Dresses-7m} dataset, and instead compare to an EdiBERT trained on this dataset and strong baselines.
% We conjecture that the GANs struggle with the strong variation exhibited in the dataset (include references). We conduct human studies on the \textit{Dresses-7m} dataset. 
% For all experiments with comparissons to EdiBERT, we use the same pre-trained first stage tokenizer for fairness.  

\input{tables/all_datasets_quant}

\subsection{Quantitative Evaluation}

\paragraph{Block edits.}

\begin{figure*}[h!]
\centering
\includegraphics[width=0.88\linewidth]{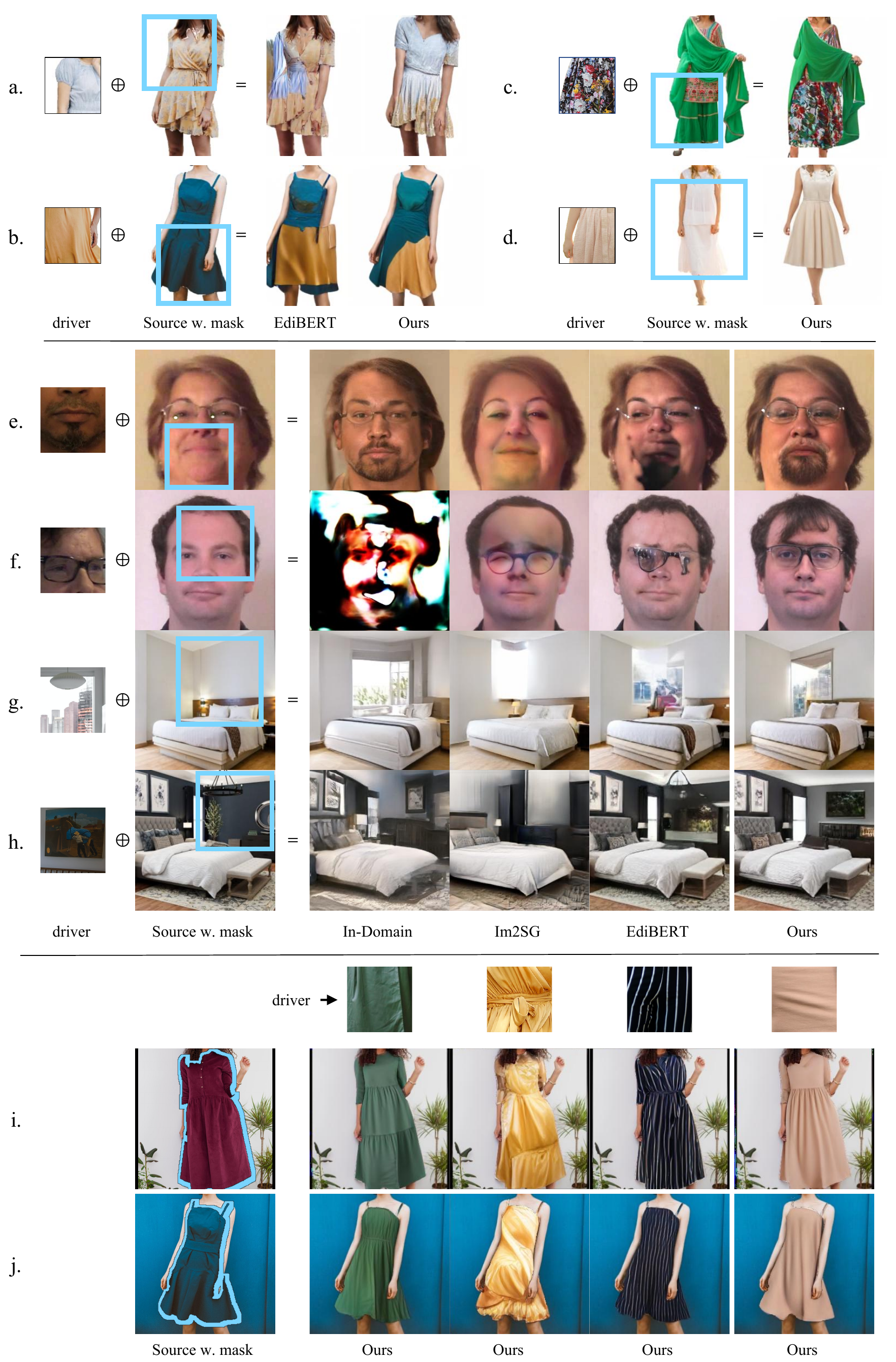}
\vspace{-3mm}
\caption{\small{Qualitative results from \method. Rows a,b,c,d: Block edits from \method trained on \textit{Dresses-7m}. Row e,f: Comparisons to prior work trained on FFHQ. Rows g,h: Comparisons to prior work trained on LSUN-Bedrooms. Rows i,k: \textit{free-form} edits from \method trained on \textit{Dresses-7m}. Please zoom in for details. In each case, the masked region in the source image is that contained within the blue line.}}\label{big_results}
\end{figure*}

\Cref{all_quant_table} reports the evaluation metrics for all baselines and datasets.
Our approach significantly outperforms others in \textbf{naturalness}:
because our method is trained explicitly with the goal of blending source and driver images, it works even for cases where the images are poorly aligned, where prior works based on fitting priors on unedited images fail (see \cref{big_results}).

The copy-paste baseline outperforms other methods on the \textbf{faithfulness} metric but has very poor naturalness --- this is expected as the edited image contains a 1-to-1 copy of the driver image.
The opposite is true for the inpainting baselines, which attain good naturalness but very poor faithfulness as they ignore the driver image altogether.
Our method is second only to copy-paste in faithfulness while also scoring best in naturalness.
Other baselines sit somewhere in between, but generally do not fair very well in faithfulness because, by projecting the composite image to the prior manifold, they distort the cue too much.

As for \textbf{locality}, copy-paste is also optimal, as it does not change the context region at all.
Compared to non-trivial baselines, our method is first or second best in this metric, affecting the context region much less than GAN methods.
However, EdiBERT is also very competitive as it is designed to leave the context nearly exactly unchanged (via \textit{periodic collage} of the output with the context).
However, relaxing locality is often necessary to obtain a more reasonable blending effect --- an intuitive fact that we show qualitatively in  \cref{big_results}f.

Finally, given that the metrics above only give an approximate idea of the quality of the edits, we also conduct a \textbf{human-study} on the \textit{Dresses-7m} dataset using Amazon Mechanical Turk.
We showed 256 edit samples using our method and EdiBERT to 3 human assessors each, asking two questions:
which of the two outputs is more realistic, and which is more faithful to the driver image.
The results show that by majority vote, human annotators think that our samples are more natural 83.2\% of the time, and more faithful 80.5\% of the time. 

\vspace{-2mm}

\paragraph{Free-form edits.}

% \AB{Here, we discuss the results for our second approach for building a dataset of ``edits'' (Section~\ref{s:data}), where a semantic segmentation network is used to ensure $x$ and $y$ are uncorrelated. 
Here, we use a semantic segmentation network to extract the edit region $R$ from image $x$,
% For evaluation data, the edit region $R$ corresponds to a semantic region in a sample image $x$, 
while we define $y$ as a crop taken from within a semantic region of a different image $x'$ in the dataset (see \cref{s:data})
We conduct experiments on \textit{Dresses-7m} and compare to EdiBERT.
We create the composited image for the EdiBERT input by tiling the driver image such that it can be pasted into $R$ in $x$.
The results are shown in \cref{all_quant_table}.
Our approach again outperforms the previous methods in terms of \textbf{naturalness} and \textbf{faithfulness} and, this time, in \textbf{locality} too, because we can better capture irregular edit regions compared to EdiBERT which results in blocky artifacts. 

\clearpage

\vspace{4mm}

\subsection{Qualitative Evaluation}

\begin{figure}[t]
\centering
\includegraphics[width=\linewidth]{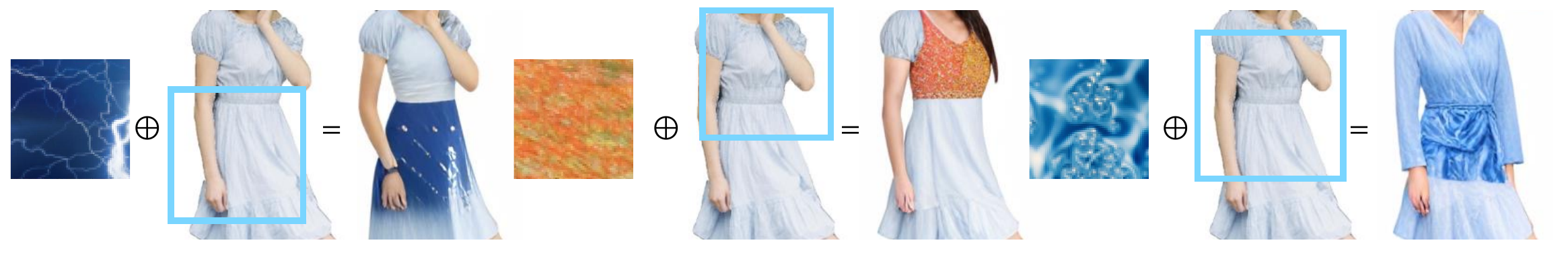}
\caption{\footnotesize{\method generalises surprisingly well to out-of-domain driver images (\textit{e.g.} images of weather and nature). Three examples using the same source image from \textit{block-edit} \textit{Dresses-7m} model. As shown, \method generates cohesive and varied samples.}}\label{OOD_targets}
\vspace{-4mm}
\end{figure}

We show qualitative comparisons against prior work in \cref{big_results}.
Our edits combine naturalness, faithfulness and locality, whereas others fail at achieving all three goals as well as we do.
% Our method is able to generate more natural images than the prior approaches.
% While previous works are unable to achieve all of of naturalness, faithfulness and locality in the generated images, our method demonstrates a combination of all three properties.
Due to the augmentations in our training edits, our method is better able to cope with uncorrelated driver images $y$ than other approaches that only rely on a pre-learned unconditional prior distribution $P(x)$.
% The previous approaches struggle when the context and driver images are not semantically aligned, as they rely on a pre-learnt unconditional prior distribution $P(x)$, which has only modelled real unedited images. 
% On the other hand, we simulate uncorrelated inputs during training with our augmentation methods, meaning that our method is more robust to non-semantically aligned inputs.
For example, in \cref{big_results}e,f our approach can successfully mix images coming from faces with different gender or pose, showing better naturalness and faithfulness.
% where inputs depicting the faces of two different sexes, or with two different poses are edited successfully by our approach. 
As for locality, while EdiBERT is highly competitive in~\cref{all_quant_table}, this comes at a cost:
in~\cref{big_results}a,f our method achieves better naturalness by coloring both sleeves in the same way and by completing the glasses even though part of them lie outside of the edit region, whereas EdiBERT cannot.
{}~\cref{big_results}i,j shows free-form edits where the entire clothing item is masked.
Although structural details of the dress are hidden by the mask, \method generates natural and varied structure that is different to the source and faithful to the driver image.
In \cref{big_results}g,h \method generates more natural looking samples than prior work that edit with respect to the spatial geometry of the room.
We also see that \method generalises surprisingly well to out-of-domain driver images, as shown in \cref{OOD_targets} for the block-edit \textit{Dresses-7m} model.
See the supp.~mat.~for additional results.

% showed that EdiBERT was highly competitive, but often relaxing locality is necessary for a more reasonable blending effect. 
% In \cref{big_results}b, our method colors both sleeves the same new color even though one lies outside of the edit region, in order to improve the cohesiveness of the edit.
% Similarly, in \cref{big_results}d, our method adds details for the glasses outside of the edit region. See the supp.~mat.~for additional results.

\vspace{-1mm}
\subsection{Ablations}
\vspace{-2mm}
In~\cref{ablation_table} we analyse and ablate design choices in \method.
We report additional metrics: negative log likelihood (NLL) on the validation set and sample diversity, computed pairwise between samples from the same inputs using LPIPS~\cite{zhang2018perceptual}.

Starting from the construction of the training edits $\mathcal{T}$ (\cref{s:data}), reducing $\alpha$ (rows a-d) means removing more of the image $\hat{x}$ from the crop $y$.
As predicted in~\cref{s:data}, removing information from $y$ increases naturalness (lower FID) at the expense of weaker faithfulness (lower R).
$\alpha=0.6$ provides a balance.
% As expected, this incerases nais hidden from the model during training.
% This largely leads to more natural samples, but that are less faithful to the driver. For small $\alpha$ the model learns to place the driver equally small and unrecognizable in the source image, while $\alpha=0.6$ offers a balance in the metrics.
% (1) more natural because 
% more information is lost from $\hat{x}$ to $(x,y,R$ is hidden from the model in training
% Reducing $\alpha$ (rows a-c) leads to more natural samples. Here, more information is hidden from the model during training, which in turn learns to generate non-trivial content in the edit region.
% Restricting all information from the model leads to the model ignoring the driver image (row d). 
\posaugment (row e vs.~b) increases naturalness by preventing the model from simply pasting the driver image in the centre of the edit region.
% This leads to significantly increased naturalness of the images in the edit-region.
% Our final model (row h) adds 
\sizeaugment (row h) randomizes the choice of $\alpha$ in a range during training, so that the editor learns to automatically resize the driver image as needed.
%  which pushes the model to choose the optimal size  at which to place the driver in the edited images by choosing $\alpha$ from a range during training.
This significantly improves naturalness at the cost of a reduction of faithfulness (row h vs.~e).
In part, this is likely due to limitations of the retrieval model used to measure faithfulness, which struggles to cope with geometric deformations even when they preserve the style of the driver.
% Although this is a worthwhile cost for the improved naturalness, we also hypothesize that the discriminative model for computing faithfulness struggles to accurately retrieve the augmented versions of the driver image that the final model generates.
$\alpha$ and augmentation have no effect on the locality (rows a-h).
A benefit is that our final model generates more diverse samples (row h vs g) by learning to place the driver image at different positions and sizes.
Interestingly, NLL is also minimised by the final model despite the fact that $y$ is \emph{less} correlated to $\hat x$ than in other cases: this is likely because additional augmentations reduce overfitting to the training data.

% We conjecture that the increased augmentation methods prevent the model from over-fitting, particularly on the smaller datasets.

Not filtering the samples by similarity to the driver (rows i,j,k) reduces faithfulness of both \method and EdiBERT (rows l,m); even so, \method outperforms all prior work in this metric (see~\cref{all_quant_table}).
Finally, training just one VQ-GAN rather than two (row n) results in a drop in faithfulness, as the single VQ-GAN struggles to faithfully reconstruct the details in the smaller driver image.

\input{tables/ablation}

%% file: tables/all_datasets_quant.tex
\begin{table}[t]
\centering
\resizebox{0.9\linewidth}{!}{
\begin{tabular}{m{2.8cm}l@{\hskip 0.4in}ll@{\hskip 0.25in}lll@{\hskip 0.15in}c}
\toprule
         &           & \multicolumn{2}{c}{Naturalness (\colorbox{lavenderblue}{$\downarrow$} )}                 & \multicolumn{3}{c}{Faithfulness \colorbox{mistyrose}{($\uparrow$)}}    & Locality (\colorbox{lavenderblue}{$\downarrow$} )    \\ \hline
            % &   & &                & \multicolumn{2}{l}{Inception distribution} & \multicolumn{2}{l}{LPIPS distance}    & \multicolumn{3}{l}{Inception features} & \multicolumn{2}{l}{L1 Outside Edit-Region}                       \\
         &            & Image             & Edit-R                & R@1         & R@5        & R@20        & (L1)                \\ \midrule

\multirow{4}{*}{\begin{tabular}[c]{@{}l@{}}\textbf{Dresses-7m} \\ (\textit{block-edits})\end{tabular}} & \light{Baseline:} Copy-Paste         &      \light{21.457}        &     \light{35.924}      &   \light{1.000}    &  \light{1.000}     &   \light{1.000}     &             \light{0.000}                                          \\
% & B: Copy-Paste VQGAN            &   -     &    -        &      \colorbox{mistyrose}{0.974}   &   \colorbox{mistyrose}{0.995}    & \colorbox{mistyrose}{1.000}     &          \colorbox{lavenderblue}{0.049}                        \\

& \light{Baseline:} Inpaint  &    \light{15.797}     &        \light{25.769}       &      \light{0.071}        &  \light{0.214}         &     \light{0.515}       &     \light{0.095}       \\
 & EdiBERT~\cite{issenhuth2021edibert}                  &    17.193     &        32.621              &   0.554         &  0.837         &     0.963       &       \textbf{0.052}                                       \\  
%  & \multicolumn{3}{l}{(ours) EdiBERT ++ }            &      &         &              &               &     &      &      &                            \\
  & (ours) \method                &   \textbf{14.411}   & \textbf{24.743}        &      \textbf{0.797} & \textbf{0.937}     & \textbf{0.978}     & 0.056              \\\midrule
  
  \multirow{7}{*}{\begin{tabular}[c]{@{}l@{}}\textbf{FFHQ}\\ (\textit{block-edits})\end{tabular}} & \light{Baseline:} Copy-Paste        &  \light{33.330}  & \light{ 25.811}          &    \light{1.000} & \light{1.000}    & \light{1.000}     &           \light{0.000}                                   \\
% & B: Copy-Paste VQGAN           &  23.812    &   32.747     &    0.501    &  0.742    &  0.905   &  0.089                   \\

& \light{Baseline:} Inpaint   &    \light{18.328}   &  \light{12.665}           &     \light{0.421}        &    \light{0.704} &    \light{0.895}    & \light{0.139}      \\
 & GAN inv~\cite{Im2SG}: StyleGANv2               &    26.583         &     16.223            &     0.590      &    0.823      &    0.948       &   0.198                           \\  
  & GAN inv~\cite{Im2SG}: StyleGANv2-Ada                        &     26.657       &         16.290        &     0.593      &    0.821      &    0.949       &  0.199                                   \\  
  & In-domain~\cite{zhu2020indomain}                  &   19.880    &      14.270              &      0.539     &     0.800     & 0.938          &            0.178                                            \\  
    % & \multicolumn{3}{l}{Latent-Regress~\cite{chai2021latent}}                  &        &           &             &                 &           &          &           &                                            &                                       \\  
 & EdiBERT~\cite{issenhuth2021edibert}                 &    13.192    &     12.230      &   0.718          &    0.925                & 0.983          &  \textbf{0.093}                                                                       \\  
%  & \multicolumn{3}{l}{(ours) EdiBERT ++ }            &      &         &              &               &     &      &      &                              \\
  & (ours) \method               &  \textbf{12.770}    &  \textbf{10.574}        &      \textbf{0.853}   &  \textbf{0.970}    &  \textbf{0.994} & 0.106                \\\midrule
    \multirow{6}{*}{\begin{tabular}[c]{@{}l@{}}\textbf{LSUN}\\\textbf{Bedrooms}\\ (\textit{block-edits})\end{tabular}} & \light{Baseline:} Copy-Paste              &     \light{24.402}     &  \light{28.828}           &    \light{1.000}  &            \light{1.000}                            &  \light{1.000}   & \light{0.000}              \\
% & B: Copy-Paste VQGAN                 &    16.965    &     24.513        &  0.628   &        0.800            & 0.915     &  0.106                       \\

& \light{Baseline:} Inpaint   &     \light{15.080}  &      \light{21.493}       &      \light{0.113}      &    \light{0.297} &  \light{0.596} &  \light{0.161}    \\
 & GAN inv~\cite{Im2SG}: StyleGANv2                 &  23.735     &    33.530        &    0.405       &  0.689        &  0.866         &      0.259              \\  
  & In-domain~\cite{zhu2020indomain}                  &   32.333     &  43.544         &           0.171     &     0.363     &     0.608      &            0.208          \\  
 & EdiBERT~\cite{issenhuth2021edibert}                    &   16.518     &    27.528          &    0.537       &   0.816       &   0.946        &     \textbf{ 0.111}                                            \\  
%  & \multicolumn{3}{l}{(ours) EdiBERT ++ }            &      &         &              &               &     &      &      &                            \\
%   & (ours) \method (?)               & 13.782     &  22.718       & 0.514    &  0.746    & 0.898     & 0.121           
  & (ours) \method             & \textbf{14.107}     &  \textbf{22.187}       & \textbf{0.789}    &  \textbf{0.923}    & \textbf{0.981}     & 0.119           \\ \midrule 
    \multirow{4}{*}{\begin{tabular}[c]{@{}l@{}}\textbf{Dresses-7m}\\ (\textit{free-form edits})\end{tabular}} & \light{Baseline:} Copy-Paste      &    \light{23.107}    &     \light{58.259}      &   \light{0.581}       &  \light{0.700}      &     \light{0.817}             & \light{0.000}                         \\
% & B: Copy-Paste VQGAN            &   -     &    -        &      \colorbox{mistyrose}{0.974}   &   \colorbox{mistyrose}{0.995}    & \colorbox{mistyrose}{1.000}     &          \colorbox{lavenderblue}{0.049}                        \\

& \light{Baseline:} Inpaint  &     \light{13.718}     & \light{24.516} &    \light{0.193}  &    \light{0.385}  &    \light{0.659}    & \light{0.103}          \\
 & EdiBERT~\cite{issenhuth2021edibert}                &  15.277     &       27.359           &      0.650   & 0.843     &   0.937     &     0.079                                 \\  
%  & \multicolumn{3}{l}{(ours) EdiBERT ++ }            &      &         &              &               &     &      &      &                            \\
  & (ours) \method                & \textbf{14.000}     &  \textbf{25.973}      &  \textbf{0.814}   & \textbf{0.920}     &     \textbf{0.951} &  \textbf{0.072}                \\ 
  \bottomrule

\end{tabular}
} 
\vspace{1mm}
\caption{\label{all_quant_table}. \footnotesize{ Results for \textit{block edits} and \textit{free-form edits}. Naturalness is computed over both the whole image (Image), and just the edit-region (Edit-R) using FID. Faithfulness is computed via retrieval, where R@K measures whether or not the sample is retrieved in the top-k instances. Locality is measured using L1 distance outside of the edit-region.}}
\vspace{-6mm}

\end{table}

% \begin{table}[!t]
% \scriptsize
% \centering
% \begin{tabular}{lll} \toprule
%               & EdiBERT & \method \\
% Q1: Realistic &         &       \\
% Q2: Faithful  &         &    \\ \bottomrule
% \end{tabular} \label{human_study}
% \caption{\label{human_study} human study}
% \end{table}

% \begin{table}[!t]
    
%     \begin{minipage}{.2\linewidth}

%       \centering
%       \scriptsize
%         \begin{tabular}{lll} \toprule
%                       & EdiBERT & \method \\
%         Q1: Realistic &         &       \\
%         Q2: Faithful  &         &    \\ \bottomrule
%         \end{tabular} \label{human_study}
%         \caption{\label{human_study} human study}
%     \end{minipage}%
%     \begin{minipage}{.9\linewidth}
%       \centering
%         \scriptsize

%         \begin{tabular}{llllll} \toprule
%                       & EdiBERT & inv~\cite{Im2SG} & In-domain~\cite{zhu2020indomain} & \method \\
%         sample time (s) &         & 91.067  & 11.169 &  &     \\\bottomrule
%         \end{tabular}
%         \caption{compute time for a sample for each method (think ours is fastest)}
%     \end{minipage} 

% \end{table}

%% file: tables/ablation.tex
\begin{table}[t!]
\resizebox{\linewidth}{!}{
\begin{tabular}{llccccc@{\hskip 0.3in}cc@{\hskip 0.3in}lll@{\hskip 0.15in}c@{\hskip 0.15in}c@{\hskip 0.15in}c@{\hskip 0.15in}cc}
\toprule
         &   &       & &     & &        & \multicolumn{2}{c}{Naturalness \colorbox{lavenderblue}{($\downarrow$)}}                    & \multicolumn{3}{c}{Faithfulness \colorbox{mistyrose}{($\uparrow$)}}    & Locality \colorbox{lavenderblue}{($\downarrow$)}    & \multicolumn{1}{c}{NLL \colorbox{lavenderblue}{($\downarrow$)}}  & \multicolumn{2}{c}{Diversity \colorbox{mistyrose}{($\uparrow$)}}   \\ \hline
  &        $\alpha$    & pos-aug&        size-aug         & Filter & 2VQ & Data &             Image          & Edit-R          & R@1         & R@5        & R@20        &          (L1)   &     &     Image             & Edit-R                  \\ \midrule

a. &0.8 & \xmark &   \xmark     & \checkmark       &\checkmark & $D$ &  17.241   &     30.076             &  0.882       &  0.980               & 0.996                                               &  0.056   & 2.181      &      0.135      &    0.309          \\
b. & 0.6 & \xmark &   \xmark           & \checkmark & \checkmark& $D$&  15.593   & 29.364          & 0.920        &   0.986        &      0.997    &  0.056                                              &  1.704         &  0.137          &        0.315      \\
 c. & 0.4 &\xmark &     \xmark        &\checkmark & \checkmark& $D$&   13.967  &  26.975                   &    0.811     &     0.954      &    0.988      &       0.056                                         &   1.594       &        0.139    & 0.327          \\
 d. & 0.0 & \xmark &   \xmark   & \checkmark   &\checkmark &$D$ &  15.797   &  25.769               &  0.071  & 0.214        & 0.515        &  0.095                                          &   1.537    &        0.190 & 0.419         \\
\hline
  e. & 0.6 & \checkmark &   \xmark   & \checkmark   &\checkmark &$D$ &    15.605 &  26.513                 &    0.887     &     0.980      &    0.997      &    0.056                                            & 1.518        &    0.142        & 0.338              \\
 f. & 0.5-0.6&\checkmark &     \checkmark       & \checkmark   &\checkmark & $D$  & 14.951    &     26.186      &  0.856       &    0.968        &     0.992     & 0.056                                               &     1.460       &      0.143      &    0.344                 \\
 g. &  0.4-0.7&\xmark &     \checkmark       & \checkmark  &\checkmark & $D$   &  14.589   &    27.824    &     0.864   & 0.970        &    0.992  &    0.056   &   1.494       & 0.139                  &  0.328   \\
h. &  0.4-0.7 & \checkmark &     \checkmark       & \checkmark  &\checkmark &   $D$    & 14.411    &    24.743       &    0.797     &  0.937         &     0.960     &   0.056              &   1.448              & 0.143           &       0.344                                   \\    \midrule
i. &    0.4-0.7 & \checkmark &     \checkmark       & \xmark   & \checkmark&  $D$    & 13.913    & 24.583         &   0.611      &        0.817   &  0.929        & 0.056                                               & 1.448     & 0.145           &     0.351                   \\
 j. &      0.4-0.7 & \checkmark &     \checkmark       & \xmark   & \checkmark&  $B$    &  14.347   & 22.998          &    0.636     & 0.831          &    0.929      &    0.119                                           &   2.942              &    0.287        &      0.460        \\
 k. &         0.4-0.7 & \checkmark &     \checkmark       & \xmark   & \checkmark&  $F$    &  12.636   &    10.699    &     0.723       &  0.899              & 0.976        & 0.106          & 2.392         & 0.203                                               & 0.321    &               \\  
 l. &      \multicolumn{3}{c}{EdiBERT~\cite{issenhuth2021edibert}  }        & \xmark   &  &  $B$ &   16.643   &  29.775      &    0.356        &      0.627          &    0.823     & 0.111          &    -       &           0.291                                     & 0.575                  \\
 m. &         \multicolumn{3}{c}{EdiBERT~\cite{issenhuth2021edibert}  }      & \xmark   &  &  $F$    &  13.036   &    12.891    & 0.536           &  0.778              &     0.925    & 0.093          &    -      &          0.181                                      &  0.423           \\  \midrule
n. &     0.4-0.7 & \checkmark &     \checkmark       & \checkmark  &\xmark &   $D$    &  14.107   &     23.916            &      0.720   & 0.891          &      0.963    &           0.056                                     &    1.454             &      0.144      &    0.347   \\       
% o. &   0.4-0.7 & \checkmark &     \checkmark       & \checkmark & \xmark&   $B$     &     &        &            &                &         &           &          &                                                &     &               \\
% p. &    0.4-0.7 & \checkmark &     \checkmark       & \checkmark   & \xmark&  $F$    &     &        &            &                &         &           &          &                                                &     &               \\
\bottomrule       
\vspace{1mm}
\end{tabular}
}
\caption{\footnotesize{Model ablations and sweeps for \textit{block edits}. Key: NLL: Negative log likelihood. $\alpha$, pos-aug, size-aug: the parameters used to define the sub-cropping transformation $T$. Filter: filtering \method samples by visual similarity to the driver image. 2VQ\@: using two VQ-GANs rather than one. Datasets: $D$: \textit{Dresses-7m}, $B$: Bedrooms, $F$-FFHQ\@. Some metrics are over both the whole image (Image), and just the edit region (Edit-R).}}\label{ablation_table}
\vspace{-5mm}
\end{table}

%% file: Sections/conclusions.tex
We have presented \method, a new approach for targeted visual image editing.
%  where a deep neural network generates a new image by modifying a region of a source image guided by a driver image.
The key innovation is an effective method for self-supervising the model end-to-end, based on only an unlabelled collection of natural images.
Using this, we can train a conditional image generator network that responds well to diverse user inputs, significantly outperforming prior work qualitatively and quantitatively despite using no manual supervision.
% our approach outperforms prior methods across qualitative, quantitative, and human analysis, and offers a new approach to learn effective image editors without supervision.

Limitations remain: our data generation technique might be difficult to extend to text-based edits
%unless dense correspondences between text and image regions can be established.
%Although not unique to our approach,
and some edits proposed by the model are unreasonable (see sup.~mat.) because the model lacks a full understanding of the semantic content of images.
Furthermore, because our model is unsupervised and data-driven, it might contain surprising unwanted biases.

% Whilst acknowledging these limitations, our approach outperforms prior methods across qualitative, quantitative, and human analysis, and offers a new approach to learn effective image editors without supervision.
Next steps include 
%upgrading the image generator to output even higher-resolution results by incorporating source image information in the output or simply training at higher resolution, and the clear extensions of 
extending \method beyond images to both spatial editing in 3D scenes and spatio-temporal editing in video. 
% and further improving the generation of training samples, for instance to account for semantic attributes automatically predicted by an image classifier.

\vspace{2mm}

\textbf{Acknowledgements.} We are grateful to Yanping Xie and Antoine Toisoul for their help with data and computing infrastructure, and to Thomas Hayes for his help with AMT. AB conducted this research during an internship at Meta AI.

%% file: Supp/Qualitative_Results.tex
\begin{figure*}[h!]
\centering
\includegraphics[width=0.97\linewidth]{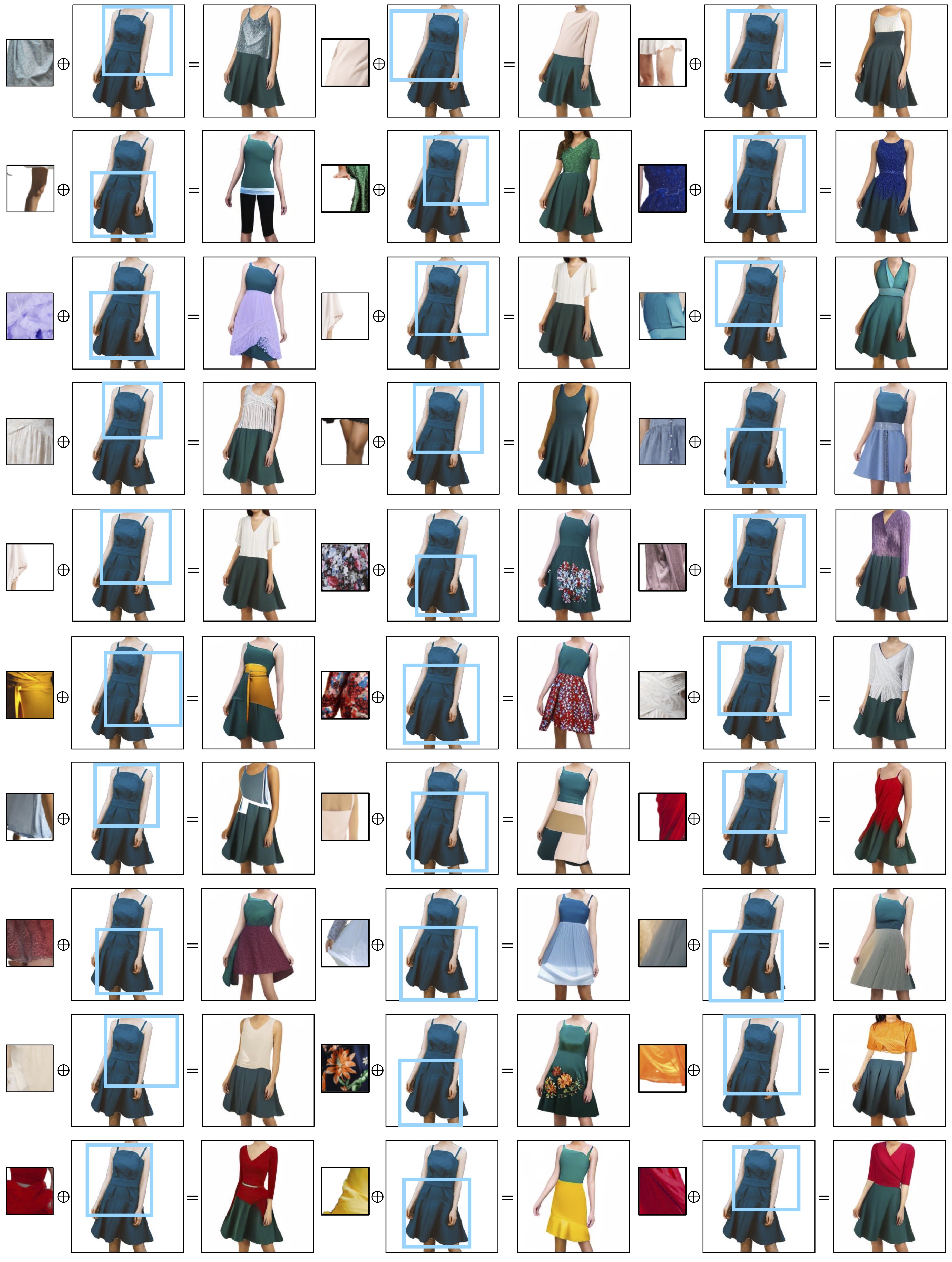}
% \vspace{-12mm}
\caption{\small{Qualitative results from the \textit{block edits} version of \method\ trained on the \textit{dresses-7m} dataset. To demonstrate the robustness of our method and the visual diversity of the generations, we fix the source image, and vary both the edit region and driver images. In each case, the masked region of the source image is that contained within the blue line. Please zoom in for details. Images are sourced from UnSplash~\cite{UnSplash}. }}\label{block_edits_dresses_sup}
\end{figure*}

In this Section we describe additional qualitative results, and analyse cases of unreasonable generations from the model. All images in this supplementary material are sourced from UnSplash~\cite{UnSplash}, or DFDC~\cite{DFDC2020}. For videos demonstrating the capabilities of \method, please see our website: \url{https://www.robots.ox.ac.uk/~abrown/E2EVE/}
% We additionally include a supplementary video demonstrating the diversity of generations for the block-edit \textit{Dresses-7m} \method\ model. 
% The video is also described in this section

\paragraph{\textit{Dresses-7m} - \textit{block edits}}. In~\cref{block_edits_dresses_sup} we show \textit{block edits} from our method on the \textit{Dresses-7m} dataset. 
To demonstrate the model robustness and sample diversity, we generate edits for the same source image, while varying the edit region and driver image. The generated edits are remarkable.
\method\ generates natural-looking edits, that are local to the edit-region and faithful to the driver images. Additionally, the edits are visually diverse, and contain many different clothing structures, styles, patterns and colors, with the same two edits rarely sharing the same generated content.
Neither the source image, nor any of the driver images in~\cref{block_edits_dresses_sup} are depicted in \textit{Dresses-7m}. However, \method\ demonstrates zero-shot generalisation capabilities to the new patterns and structures in these images in order to blend the content of the driver images cohesively with the source image.

% \paragraph{Supplementary Video}. We include a supplementary video in this material. The video shows generated \textit{block edits} from our method on the \textit{Dresses-7m} dataset. To demonstrate the diversity and robustness of the model generations, we visualise the effects of varying the edit region while keeping the source and driver images the same. \method\ generates a diverse range of clothing structures and styles depending on where the edit region is placed, which are also faithful to the driver image.

\begin{figure}[hbtp]
\centering
\includegraphics[width=\linewidth]{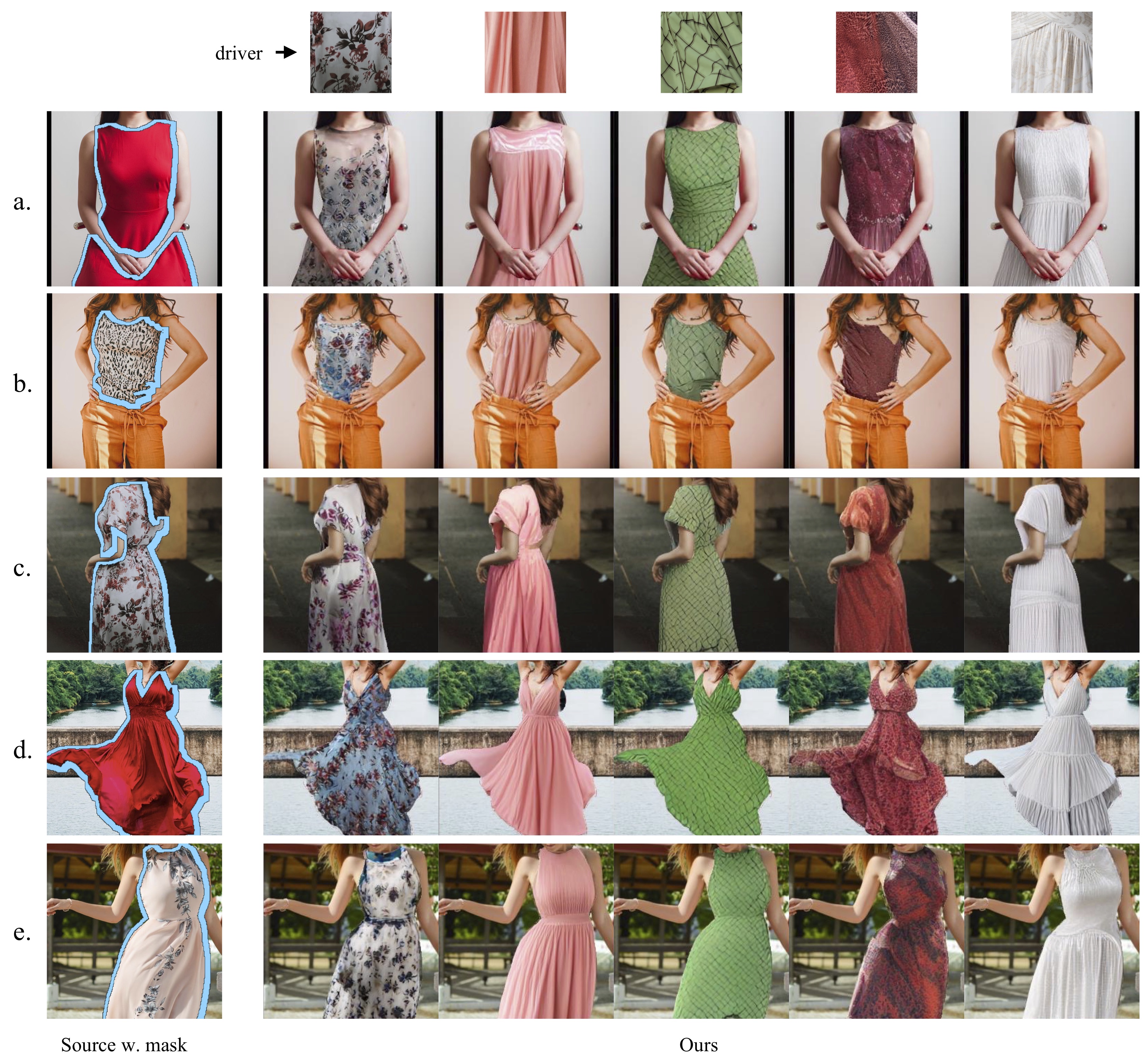}
\caption{\small{Qualitative results from the \textit{free-form} edits version of \method\ trained on the \textit{dresses-7m} dataset. We show all image generation permutations for 5 masked source images, and 5 driver images. In each case, the masked region of the source image is that contained within the blue line. Please zoom in for details. Images are sourced from UnSplash~\cite{UnSplash}.} }\label{free_form_edits_dresses_sup}
\vspace{-3mm}
\end{figure}

\paragraph{\textit{Dresses-7m} - \textit{free-form} edits}. We show additional \textit{free-form} edits from \method\ on the \textit{Dresses-7m} dataset in~\cref{free_form_edits_dresses_sup}.
% Here, the edit region $R$ is the output of a semantic segmentation network which segments the clothing item in the source image, while the driver images are taken from a region within a clothing item from a different image. 
Although all structural information besides the outline of the clothing item (\textit{i.e.} the outline of the mask) are masked in the source image, \method\ generates natural and diverse clothing structures for the same source image and edit region (\textit{e.g.} see the different waist and neckline structures in \cref{free_form_edits_dresses_sup}a), that are faithful to the driver images. 
Impressively, although the training set of \textit{Dresses-7m} contains only dresses, \method\ is able to generalise well to an edit region depicting a T-shirt in~\cref{free_form_edits_dresses_sup}b.

Because the edit region $R$ fully contains an object (in this case, a clothing item), and separates this object in the 
foreground from the background, we are hence able to paste the edited region back on to the source image, without creating any disjoint spatial continuity over the edit region boundary in the output.
This has the effect of removing any non-local effects from the edit, while preserving the naturalness of the source image outside of the edit region. 
All generated results in~\cref{free_form_edits_dresses_sup} are hence shown with the generated edit region pasted back onto the source image. 

\begin{figure}[t]
\centering
\includegraphics[width=\linewidth]{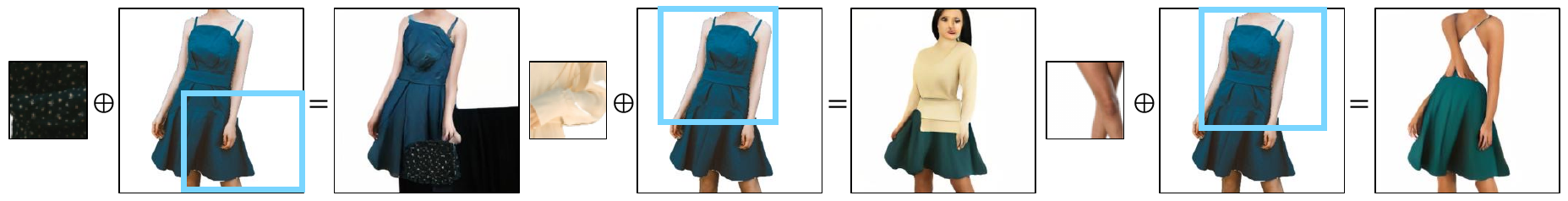}
\caption{\small{Examples of unreasonable generations from \method.
In a small number of samples from the \textit{block-edits} model trained on \textit{Dresses-7m}, some unreasonable generations are seen. Left: Part of the edit region is left in the output image. Middle: An unnatural looking face is generated occasionally when the edit region fills the top-most part of the source image. Right: Non-aligned inputs can lead to unreasonable generations.  }}\label{bad_generations}
\vspace{10mm}
\end{figure}

\paragraph{Unreasonable Generations - \method}. \method\ generates very impressive results even in cases when the driver and source images are highly uncorrelated (see Fig. 5 in main paper, and ~\cref{block_edits_dresses_sup} in the supp. mat.). However, there are still some edits proposed by the model that are unreasonable. This is not a problem unique to our approach, and is due to the model lacking a full understanding of the semantic content of images.
\cref{bad_generations} highlights three cases of unreasonable edits that were seen in a small number of generated images from  \method\ using \textit{block-edits} trained on \textit{Dresses-7m}.

\cref{bad_generations} left: \method\ leaves part of the masked edit region in the output generated image. 
\method\ simply marks the masked region in the model inputs by leaving a $R$-shaped hole in the masked source image (Section 3.2 in main paper), and very occasionally, part of this hole is left in the model output.
Interestingly, this only happens when $R$ includes the bottom-most rows of the image. 

\cref{bad_generations} middle: \method\ generates an unnatural-looking face. The \textit{Dresses-7m} dataset contains some faces. When trained on this dataset, the model hence learns that faces are likely to appear in the top-most parts of images. When the edit region includes the top-most part of the image, \method\ may generate an unnatural looking face. These generations could be avoided by removing faces from the training set.

% This is because of the rarity of faces appearing in the dataset, and also due our familiarity with human-faces.

\cref{bad_generations} right: \method\ generates an unreasonable edit when the source and driver images are completely mis-aligned. In this case \method\ is asked to blend some legs into the top of a dress. This is an impossible edit to complete naturally, and one that would not appear in the self-supervised training regime used by the method. Impressively, \method\ is still able to generate cohesive blends occasionally when given such non-aligned inputs (see top-right-most generation in~\cref{block_edits_dresses_sup}).

\begin{figure*}[h!]
\centering
\includegraphics[width=\linewidth]{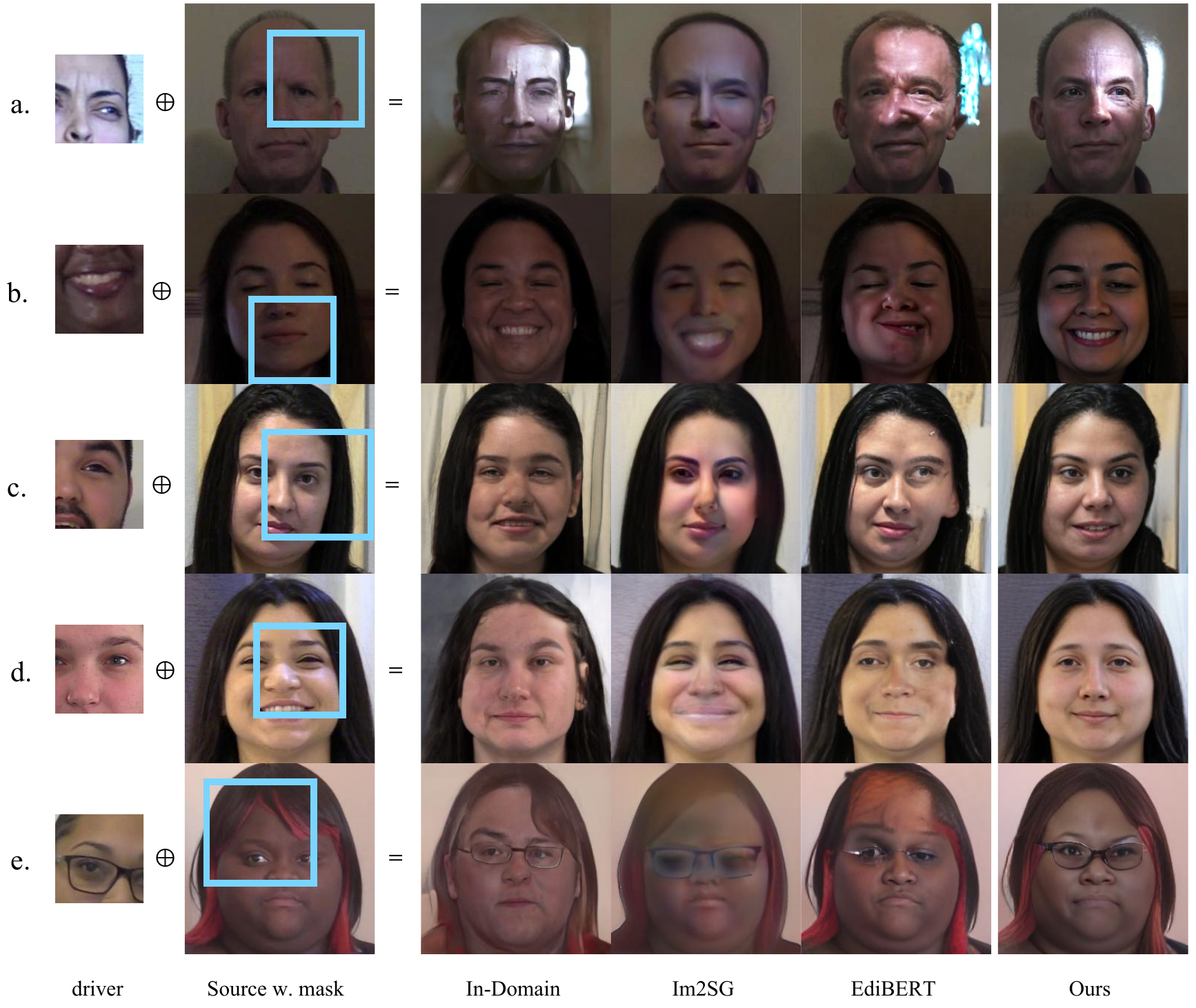}
\vspace{-3mm}
\caption{\small{Qualitative comparisons to prior work when trained on FFHQ. Our method generates edits that are natural-looking, faithful to the driver and local to the edit region, whereas prior work struggles to achieve a balance of all three. In each case, the masked region of the source image is that contained within the blue line. Images are sourced from DFDC~\cite{DFDC2020}.  }}\label{sup_face_compare}
\end{figure*}

\paragraph{FFHQ - Comparisons. } In~\cref{sup_face_compare} we show some qualitative comparisons to prior work for the models trained on the FFHQ dataset. Our method generates edits that are natural-looking, faithful to the driver, and that are local, whereas the prior work struggles to achieve all three (see Table 1 in the main paper for the quantitative demonstration of this trend).

\begin{figure*}[h!]
\centering
\includegraphics[width=0.81\linewidth]{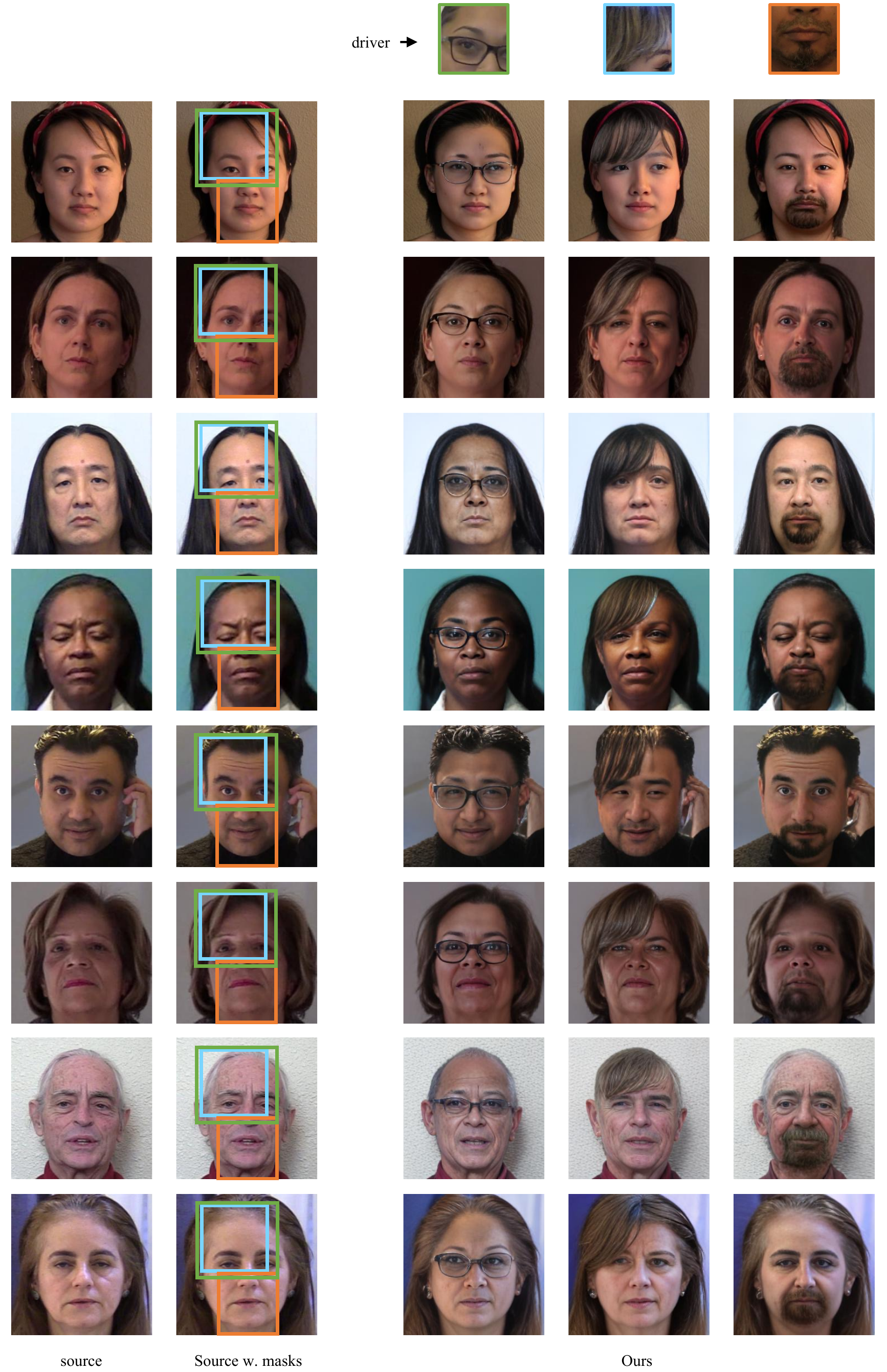}
\vspace{-3mm}
\caption{\small{Qualitative results for our method when trained on FFHQ. For each source image (row) we use the same three driver image and edit region pairs to generate three edited images. To save space, we display the three different edit regions used for the three samples on the same image (the \textit{source w. masks} column). The left column of generated images corresponds to the green edit region, the middle column corresponds to the blue edit region, and the right column corresponds to the orange region. This figure is best viewed in color. Images are sourced from DFDC~\cite{DFDC2020}. }}\label{sup_face_permute}
\end{figure*}

\paragraph{FFHQ - \method. } In~\cref{sup_face_permute} we show additional qualitative results from our method trained on FFHQ. Here, we probe whether \method\ is able to make different edits to the same source image, or make the same edit to many different source images. Specifically, we choose three driver image and edit region pairs, $(y,R)$, and pair them with a number of varied source images. 

\method\ generates natural-looking edits, even when the driver and source images are non-aligned (\textit{e.g.} due to the different genders in the source and driver images).
Impressively the generated images are faithful to the specific details of the driver image across all of the source images. For example, the style of the edited glasses and new hairstyle specifically match the corresponding driver images.
This highlights the capability of image-based visual editing to explicitly edit specific visual content.
Furthermore \method\ makes stylistic changes to the driver image to make for a more natural edit that is cohesive with the surrounding source image. For example, the added hair (both head and beard) are manipulated in color and texture to roughly match the rest of the hair in the image.

% (4) Occasionally a lot of the face changes - but that is because that part of the image was within the mask - so can just make it smaller

\clearpage

\clearpage

\clearpage

\clearpage

% \subsection{Image Driven is better than text driven}
% Here, we analyse our claim that image driven manipulation is better than text driven manipulation. For this, I will use the ``Paint-by-word'' method, and try to describe some image edits with text for the bedrooms dataset

%% file: Supp/Implementation_Details.tex
In this Section, we provide further details on how our method is implemented. Specifically we detail how the VQ-GANs that we use in our method are trained. This is in addition to Section 3.3 in the main paper. We also give further clarification on how triplets are formed for the evaluation data. This is in addition to Section 4 in the main paper.

\subsection{VQ-GAN Training} \label{VQ_gan_training_sup}
Here, we explain the loss function for training the VQ-GAN.
Following~\cite{Esser2021TamingTF}, the VQ-GAN is trained to optimize the following loss function:

\begin{equation}
\mathrm{L}_{VQ}(E,G,Z) =  \norm{x - \hat{x}}^2 + \norm{sg[\Phi(x)] - z}^{2}_{2} + \norm{sg[z] - \Phi(\hat x)}^{2}_{2}
\label{VQ_GAN_loss}
\end{equation}

where $x$ is the input image, $\hat{x} = \Psi(z)$ is the reconstructed image, $z = \Phi(x)$ are the discrete codes, $\Phi$ is the encoder, and $\Psi$ is the paired decoder. The first term is a reconstruction loss, the second a loss pulling the codebook vectors towards the encoder outputs, and the third term is the commitment loss~\cite{vq_vae}, which makes sure that the encoder commits to an embedding and the output does not grow arbitrarily. Here, $sg[\cdot]$ is the stop-gradient operation. Backpropogation through the non-differentiable quantization step is computed via a straight through gradient estimator. For more details and motivation see~\cite{Esser2021TamingTF,Ramesh2021ZeroShotTG,Salimans2017PixeCNN,vq_vae}. 

\subsection{Constructing Evaluation Data Triplets}
Here, we give further clarification on how evaluation data triplets are formed. Following Section 4 in the main paper, each triplet $(x,y,R)$ in the evaluation data consists of the source image $x$, the edit region $R$, and the driver image $y = x'|_R$ that is taken from the same spatial location (\textit{i.e.} the coordinate centres of both driver image and edit region align), but from a different image $x'$. We add the further detail that the edit region $R$ is purposefully chosen to be larger than the driver image $y$. Due to the lack of strict alignment in the datasets (apart from FFHQ), this rule ensures that there is a high chance of there existing a feasible edit for each triplet.

%% file: Supp/Baseline_Implementation_Details.tex
In this Section, we provide further details on how the baselines that we compare to in our experiments are implemented. We explain how we form the inputs for the baselines in~\cref{forming_composited_inputs}, we list the hyper-parameters and code sources for the baseline methods in~\cref{baseline_implement_sup}, and we analyse the quantitative effect of certain baseline implementation design choices in~\cref{quant_analyse_baselines}.

\subsection{Constructing Inputs for Baseline methods} \label{forming_composited_inputs}
Here, we detail how we form the inputs for each of the prior work baselines. Whereas our method takes as input the driver image $y$ and masked source image $x$ as separate pieces of conditioning information, the prior work baselines instead take as input a composited image, where the driver image has been pasted onto the source image. In the corresponding papers for each of the baseline approaches, the composited image is constructed such that the driver and source images semantically align in terms of both scale and positioning. This is done either manually, or by relying on the strict alignment of datasets such as FFHQ. However, in our work, we are not limited to aligned datasets, and wish to avoid manual intervention when forming evaluation data. A challenge hence exists in how to construct the composited images automatically from the evaluation data triplets $(x,y,R)$. 

Recall that for quantitative evaluation, we generate 10 edits from each method for each evaluation data triplet. To this end, for each evaluation data triplet, we form 10 composited images for each evaluation data triplet by pasting the driver image at different positions within the edit region $R$ in $x$. The edit region $R$ is larger than the driver image $y$ so simply pasting $y$ into $R$ in $x$ without resizing would still leave a hole in $x$ within $R$ around the driver $y$. This would unfairly disadvantage the prior work, as such a hole would place the composited image out of the domain of natural images that the prior work models have been trained on. Instead, we inpaint the remaining hole with the underlying image content from $x$. This should instead have the opposite effect of advantaging the prior work baselines over our method, seeing as they are shown more of the source image $x$ in their input.

\subsection{Baseline Implementation Details} \label{baseline_implement_sup}
Here, we detail how the baseline methods that are compared to in Section 4 in the main paper were implemented. Where possible, we use official code repositories, and used the default hyper-parameters recommended by the authors. For the cases where we used hyper-parameters different to those recommended by the authors, we have provided quantitative analysis justifying these choices in~\cref{quant_analyse_baselines}.

\paragraph{EdiBERT~\cite{issenhuth2021edibert}.}
We use the official code repository and follow the author's guidance for their \textit{image composition} experimental setting. Namely, we dilate the mask by 1 token to reduce border irregularities, we periodically collage the image with the input, and use spiral ordering for sampling edited tokens. An additional parameter is the number of optimisation epochs. In each optimisation epoch, all tokens in the edit region are updated once. Although the authors set the number of optimisation epochs to 2 for image composition, we find empirically that using 1 epoch obtains a better balance between metrics, and these are the results that we report in the main paper.
% ~\cite{edibertgit}

\paragraph{GAN inv~\cite{Im2SG}.}

We use the official code repository of StyleGAN2-Ada~\cite{Karras2020ada} for the implementation of GAN inv~\cite{Im2SG}. We use the default recommended optimization hyper-parameters for projecting given images into a pretrained GAN latent space.
% ~\cite{stgv2adagit}
\paragraph{In-Domain~\cite{zhu2020indomain}.} 

We use the official code repository, and follow their default implementation for \textit{Semantic Diffusion}. The In-Domain approach follows a two stage pipeline, with a GAN inversion stage, followed by a domain-regularised optimisation stage.
We find empirically that removing the regularisation stage results in a better balance between metrics, and these are the results that we report in the main paper. 
% ~\cite{indomaingit}

\input{tables/supp_table_extra}

\subsection{Quantitative Analysis of Baseline Design Choices} \label{quant_analyse_baselines}

Here, we analyse the quantitative effect of two baseline implementation design choices. Specifically, we analyse the effect of the number of optimisation epochs in EdiBERT, termed \textit{EdiBERT~\cite{issenhuth2021edibert} (n epoch)}, where \textit{n} refers to the number of optimisation epochs. 
We also analyse the effect of either including the regularisation stage in the In-Domain method (termed, \textit{ In-domain~\cite{zhu2020indomain} w. reg}), or not (termed, \textit{ In-domain~\cite{zhu2020indomain}}). 
We use the same metrics as used in Section 4 in the main paper, namely, naturalness, faithfulness and locality. The results are shown in~\cref{sup_table_quant}.

For EdiBERT, increasing the number of optimisation epochs results in more natural-looking samples, but this is at the cost of a sharp drop in faithfulness. 
This is as expected because the EdiBERT optimization procedure improves the likelihood of the generated image with respect to the learnt unconditional image prior, with little constraint in keeping faithfulness to the driver. 
We choose to report numbers for 1 epoch of optimization in the main paper, as this offers the most competitive balance between metrics. 
For all versions of EdiBERT, our method is still superior in terms of naturalness and faithfulness, as reported in the main paper.

For In-Domain, adding the regularisation means that the model becomes far more faithful to the driver image, but at the cost of a large drop in locality and naturalness. In fact, for the FFHQ dataset, the regularisation method outperforms \method\ in terms of faithfulness, but this is at the cost of the source image being no longer recognizable with poor naturalness and locality of 0.321. Hence, we report metrics in the main paper for In-Domain without the regularisation, as only in this version of the method where the editing can be considered local.

%% file: tables/supp_table_extra.tex
\begin{table}[t]
\centering
\resizebox{0.9\linewidth}{!}{
\begin{tabular}{m{2.8cm}l@{\hskip 0.4in}ll@{\hskip 0.25in}lll@{\hskip 0.15in}c}
\toprule
         &           & \multicolumn{2}{c}{Naturalness (\colorbox{lavenderblue}{$\downarrow$} )}                 & \multicolumn{3}{c}{Faithfulness \colorbox{mistyrose}{($\uparrow$)}}    & Locality (\colorbox{lavenderblue}{$\downarrow$} )    \\ \hline
            % &   & &                & \multicolumn{2}{l}{Inception distribution} & \multicolumn{2}{l}{LPIPS distance}    & \multicolumn{3}{l}{Inception features} & \multicolumn{2}{l}{L1 Outside Edit-Region}                       \\
         &            & Image             & Edit-R                & R@1         & R@5        & R@20        & (L1)                \\ \midrule

% & B: Copy-Paste VQGAN            &   -     &    -        &      \colorbox{mistyrose}{0.974}   &   \colorbox{mistyrose}{0.995}    & \colorbox{mistyrose}{1.000}     &          \colorbox{lavenderblue}{0.049}                        \\
\multirow{6}{*}{\begin{tabular}[c]{@{}l@{}}\textbf{Dresses-7m} \\ (\textit{block-edits})\end{tabular}} & \light{Baseline:} Copy-Paste         &      \light{21.457}        &     \light{35.924}      &   \light{1.000}    &  \light{1.000}     &   \light{1.000}     &             \light{0.000}                                          \\
% & B: Copy-Paste VQGAN            &   -     &    -        &      \colorbox{mistyrose}{0.974}   &   \colorbox{mistyrose}{0.995}    & \colorbox{mistyrose}{1.000}     &          \colorbox{lavenderblue}{0.049}                        \\

& \light{Baseline:} Inpaint  &    \light{15.797}     &        \light{25.769}       &      \light{0.071}        &  \light{0.214}         &     \light{0.515}       &     \light{0.095}       \\
 & EdiBERT~\cite{issenhuth2021edibert}  (1 epoch)  $\star$              &    17.193     &        32.621              &   0.554         &  0.837         &     0.963       &       \textbf{0.052}                                       \\  

    & EdiBERT~\cite{issenhuth2021edibert}  (2 epoch) $\dagger$               &   16.058     &  31.800                & 0.404        &  0.711       & 0.923          &  \textbf{0.052}                                         \\  
  & EdiBERT~\cite{issenhuth2021edibert} (3 epoch)                 &  15.570      &     31.727            &        0.325  &     0.625    &  0.890        &  \textbf{0.052}                                 \\  
%  & \multicolumn{3}{l}{(ours) EdiBERT ++ }            &      &         &              &               &     &      &      &                            \\
  & (ours) \method                &   \textbf{14.411}   & \textbf{24.743}        &      \textbf{0.797} & \textbf{0.937}     & \textbf{0.978}     & 0.056              \\\midrule

% & B: Copy-Paste VQGAN           &  23.812    &   32.747     &    0.501    &  0.742    &  0.905   &  0.089                   \\
  \multirow{8}{*}{\begin{tabular}[c]{@{}l@{}}\textbf{FFHQ}\\ (\textit{block-edits})\end{tabular}} & \light{Baseline:} Copy-Paste        &  \light{33.330}  & \light{ 25.811}          &    \light{1.000} & \light{1.000}    & \light{1.000}     &           \light{0.000}                                   \\
% & B: Copy-Paste VQGAN           &  23.812    &   32.747     &    0.501    &  0.742    &  0.905   &  0.089                   \\

& \light{Baseline:} Inpaint   &    \light{18.328}   &  \light{12.665}           &     \light{0.421}        &    \light{0.704} &    \light{0.895}    & \light{0.139}      \\
& In-domain~\cite{zhu2020indomain}       $\star$           &   19.880    &      14.270              &      0.539     &     0.800     & 0.938          &            0.178                                            \\  
    & In-domain~\cite{zhu2020indomain} w. reg    $\dagger$              & 24.192     &     13.733           &  \textbf{0.953}     & \textbf{0.988}     &     \textbf{0.995}  &            0.321                                       \\  
    % & \multicolumn{3}{l}{Latent-Regress~\cite{chai2021latent}}                  &        &           &             &                 &           &          &           &                                            &                                       \\  
 & EdiBERT~\cite{issenhuth2021edibert} (1 epoch)  $\star$              &    13.192    &     12.230      &   0.718          &    0.925                & 0.983          &  0.093                                                                       \\  
  & EdiBERT~\cite{issenhuth2021edibert}  (2 epoch)  $\dagger$              & 13.450       &  10.874                & 0.675        &  0.895       &  0.976         &      \textbf{0.092}                                     \\  
  & EdiBERT~\cite{issenhuth2021edibert} (3 epoch)                 &  13.496      &     10.739            & 0.640         &  0.870        &  0.970        &             \textbf{0.092}                      \\  
%  & \multicolumn{3}{l}{(ours) EdiBERT ++ }            &      &         &              &               &     &      &      &                              \\
  & (ours) \method               &  \textbf{12.770}    &  \textbf{10.574}        &      0.853   &  0.970    &  0.994 & 0.106                \\\midrule

    \multirow{6}{*}{\begin{tabular}[c]{@{}l@{}}\textbf{LSUN}\\\textbf{Bedrooms}\\ (\textit{block-edits})\end{tabular}} & \light{Baseline:} Copy-Paste              &     \light{24.402}     &  \light{28.828}           &    \light{1.000}  &            \light{1.000}                            &  \light{1.000}   & \light{0.000}              \\
    & \light{Baseline:} Inpaint   &     \light{15.080}  &      \light{21.493}       &      \light{0.113}      &    \light{0.297} &  \light{0.596} &  \light{0.161}    \\
% & B: Copy-Paste VQGAN                 &    16.965    &     24.513        &  0.628   &        0.800            & 0.915     &  0.106                       \\

  & In-domain~\cite{zhu2020indomain}      $\star$            &   32.333     &  43.544         &           0.171     &     0.363     &     0.608      &            0.208          \\  
      & In-domain~\cite{zhu2020indomain} w. reg  $\dagger$                &   46.566   & 42.718               &    0.677   & 0.815     &  0.914     &  0.326                                                 \\  
 & EdiBERT~\cite{issenhuth2021edibert}     (1 epoch) $\star$              &   16.518     &    27.528          &    0.537       &   0.816       &   0.946        &     \textbf{0.111}                                           \\  
  & EdiBERT~\cite{issenhuth2021edibert}  (2 epoch)$\dagger$                 &    15.791    & 26.234                 & 0.392         &    0.712     & 0.903          &   \textbf{0.111}                                    \\  
  & EdiBERT~\cite{issenhuth2021edibert} (3 epoch)                 &   15.696     & 27.384                &     0.316     & 0.629        & 0.871        & \textbf{0.111 }                                 \\ 
%  & \multicolumn{3}{l}{(ours) EdiBERT ++ }            &      &         &              &               &     &      &      &                            \\
%   & (ours) \method (?)               & 13.782     &  22.718       & 0.514    &  0.746    & 0.898     & 0.121           
  & (ours) \method             & \textbf{14.107}     &  \textbf{22.187}       & \textbf{0.789}    &  \textbf{0.923}    & \textbf{0.981}     & 0.119           \\             
  \bottomrule

\end{tabular}
} 
\vspace{1mm} 
\caption{\label{sup_table_quant} \footnotesize{Results for \textit{block-edits} when analysing design choices for baseline implementations. When implementing prior work, we use default recommended implementation settings where possible. However, we find that different implementation settings for In-Domain and EdiBERT lead to a more preferable balance of the metrics. Here, we analyse the effect of these implementation details. Key: $\dagger$ refers to the design choice recommended by the authors of the respective paper. $\star$ refers to the design choice that we report numbers for in the main paper. Results for our method and the simple baselines have been included for ease of reference.}}
\end{table}

%% file: Supp/Quantitative_Results.tex
In this Section we provide additional quantitative results. These results explore the computational efficiency of our approach compared to prior work (\cref{comp_efficient_section}), and different sampling techniques (\cref{sampling_analysis})

\subsection{Computational Efficiency of \method} \label{comp_efficient_section}
The throughput (measured in generated images/second) for \method\ compared to each baseline method is shown in~\cref{computational_cost_table}. 

\input{tables/timing}

\method\ notably achieves higher throughput than some GAN-based approaches (GAN inv, and In-domain w. reg). Because \method\ is trained end-to-end for the editing task, generated images can be sampled directly from the model. This avoids the costly test-time optimisation processes necessary for these GAN-based approaches. 

Additionally, \method\ achieves comparable throughput with the attention-based baseline EdiBERT, and even achieves higher throughput when EdiBERT uses more than 1 optimisation epoch (the authors recommend using 2). Whereas \method\ samples every token of the output generated image during inference, EdiBERT only samples the tokens within the edit region. This means that EdiBERT can achieve higher throughput by requiring less sampling iterations, but this is at the cost of not making any non-local edits that may be necessary for improving the overall naturalness of the generated image (see section 4.2 in main paper). Although \method\ samples more tokens than EdiBERT during inference, \method\ uses a significantly smaller backbone transformer (24 vs 32 layers), leading to comparable throughput times.

In-domain achieves very competitive throughput when not using the regularisation stage (\textit{In-domain~\cite{zhu2020indomain}}). This speed is expected from and is an advantage of the simple encoder-decoder inversion architecture. However, this fast inference speed comes at the cost of significantly worse generated samples than \method\ across naturalness, faithfulness and locality (see Table 1 in the main paper and Section~\ref{quant_analyse_baselines}).

\subsection{Effect of Sampling Methods} \label{sampling_analysis}

\begin{figure*}[t]
\centering
\includegraphics[width=\linewidth]{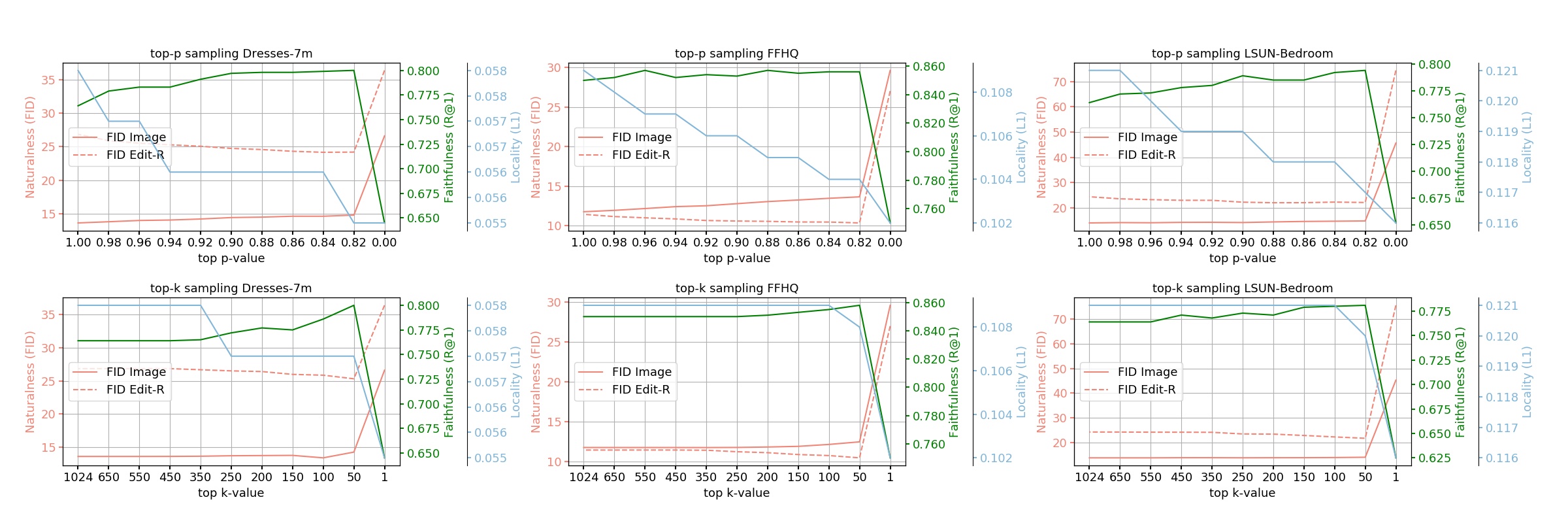}
\vspace{-3mm}
\caption{\small{The effect of different sampling methods on the naturalness, faithfulness and locality metrics across different datasets. We analyse the effect of the p-value and k-value for top-p and top-k sampling methods, respectively. }}\label{sampling_analysis_figure}
\end{figure*}

We explore the effect of different sampling methods in~\cref{sampling_analysis_figure} on the faithfulness, naturalness and locality metrics across all datasets. At each sampling step during inference, a token is sampled from the output probability histogram from the model. Here, we analyse two different sampling techniques: first, top-k sampling, where the probabilities are first sorted, and only the top-k are sampled from. Second, top-p sampling (\textit{nucleus} sampling~\cite{Holtzman2020The}), where the probabilities are sorted, and only those with a cumulative probability less than the p-value are sampled from. 

A top-p value of 0 and a top-k value of 1 results in a deterministic (or \textit{greedy}) sampling process where the most likely token is sampled at each step. A top-p value of 1 and a top-k value of 1024 means that every token is considered at each step. Several interesting conclusions can be made.

First, deterministic sampling leads to a sharp drop in naturalness and faithfulness. This is an expected result, as simply choosing the highest probability token at each step tends to not result in the most probable sequences.

Second, aside from deterministic sampling, the performance across all metrics and datasets is fairly consistent and robust across all top-p and top-k values, as indicated by the small range that the metrics change over outside of a top-p value of 0 and top-k value of 1.
This is in contrast to~\cite{Esser2021TamingTF} where naturalness (FID) was observed to severely degrade for unconditional generation when all tokens are sampled over.
This indicates that in our case the distribution at each sampling step is narrow/peaked, meaning that there are just a few tokens with high probability that can be sampled from reasonably. Hence, the most likely tokens hold so much of the probability mass that considering the long tail does not affect the generated images drastically.
This makes sense, because rather than generating images unconditionally, \method\ is generating edited images, where the content of the output image is often easily predicted from the conditioning information. 

Third, locality is optimal for deterministic/greedy sampling and degrades once more tokens are considered in the sampling.
To explain this, we note that simply copying the source image outside of the edit region would lead to the best performance in the locality metric. 
We conjecture that the most probable token at each step outside the edit region corresponds to the spatially corresponding token in the source image.
By observing the probability histograms outside of the edit region, we see that often there is one token that takes almost all of the probability mass, and this likely corresponds to the corresponding token in the source image. 
When more tokens are considered during sampling, occasionally non-local edits are sampled from the long tail of the histogram.

%% file: tables/timing.tex
\begin{table}[t]
\scriptsize
\centering
\begin{tabular}{l@{\hskip 0.4in}c} \toprule
    Method     & \begin{tabular}[c]{@{}c@{}}Throughput  \colorbox{mistyrose}{($\uparrow$)}\\ img/s\end{tabular} \\ \midrule
GAN inv~\cite{Im2SG}: StyleGANv2 &      00.01      \\
GAN inv~\cite{Im2SG}: StyleGANv2-Ada &   00.01      \\
 In-domain~\cite{zhu2020indomain} w. reg &    00.17       \\
  EdiBERT~\cite{issenhuth2021edibert}     (3 epoch)&  00.23           \\
  EdiBERT~\cite{issenhuth2021edibert}  (2 epoch) &   00.33   \\ 
 EdiBERT~\cite{issenhuth2021edibert}     (1 epoch)&    00.67    \\
In-domain~\cite{zhu2020indomain}    &        33.33     \\ \midrule

(ours) \method &   00.27     \\\bottomrule
\end{tabular} 
\vspace{2mm}
\caption{\label{computational_cost_table} Computational efficiency of \method\ compared to baseline approaches as measured by throughput (generated images per second). Throughput is computed via time taken to generate a single sample with batch size of 1 on an NVIDIA A100. throughput is averaged over multiple samples.}
\end{table}